\documentclass[journal]{IEEEtran}
\usepackage{cite}
\usepackage{amsmath,graphicx}
\usepackage{epstopdf}
\usepackage{xcolor}
\usepackage{amssymb}
\usepackage{pifont}
\usepackage{array}
\usepackage{booktabs}
\usepackage{bm}
\usepackage{multirow}
\usepackage{graphicx}
\usepackage{color}
\usepackage{float}
\usepackage{stfloats}
\usepackage[misc]{ifsym}
\usepackage{stmaryrd}
\usepackage{grffile}
\usepackage{subfig}
\usepackage{makecell}
\usepackage{threeparttable}
\usepackage{enumitem}
\usepackage{enumerate}
\usepackage[colorlinks,linkcolor=red]{hyperref}

\begin{document}
 \UseRawInputEncoding 
\title{Arbitrary-Oriented Ship Detection through Center-Head Point Extraction}
\author{Feng Zhang, Xueying Wang, Shilin Zhou, Yingqian Wang, Yi Hou

\thanks{This work was partially supported in part by the National Natural Science Foundation of China (Nos. 61903373, 61401474, 61921001).}

\thanks{Feng~Zhang, Xueying~Wang, Shilin~Zhou, Yingqian~Wang, Yi~Hou are with the College of Electronic Science and Technology, National University of Defense Technology (NUDT), P. R. China. Emails: \{zhangfeng01, wangxueying, slzhou, wangyingqian16, yihou\}@nudt.edu.cn. (Corresponding author: Xueying~Wang)}}

\markboth{Submitted to IEEE Transactions on Geoscience and Remote Sensing}%
{Shell \MakeLowercase{\textit{et al.}}: Bare Demo of IEEEtran.cls for IEEE Journals}

\maketitle

\begin{abstract}
 Ship detection in remote sensing images plays a crucial role in various applications and has drawn increasing attention in recent years. However, existing arbitrary-oriented ship detection methods are generally developed on a set of predefined rotated anchor boxes. These predefined boxes not only lead to inaccurate angle predictions but also introduce extra hyper-parameters and high computational cost. Moreover, the prior knowledge of ship size has not been fully exploited by existing methods, which hinders the improvement of their detection accuracy. Aiming at solving the above issues, in this paper, we propose a \emph{center-head point extraction based detector} (named CHPDet) to achieve arbitrary-oriented ship detection in remote sensing images. Our CHPDet formulates arbitrary-oriented ships as rotated boxes with head points which are used to determine the direction. And rotated Gaussian kernel is used to map the annotations into target heatmaps. Keypoint estimation is performed to find the center of ships. Then, the size and head point of the ships are regressed. The orientation-invariant model (OIM) is also used to produce orientation-invariant feature maps. Finally, we use the target size as prior to finetune the results. Moreover, we introduce a new dataset for multi-class arbitrary-oriented ship detection in remote sensing images at a fixed ground sample distance (GSD) which is named FGSD2021. Experimental results on FGSD2021 and two other widely used data sets, i.e., HRSC2016, and UCAS-AOD demonstrate that our CHPDet achieves state-of-the-art performance and can well distinguish between bow and stern. Code and FGSD2021 dataset are available at
 \url{https://github.com/zf020114/CHPDet}.
 \end{abstract}
 \begin{figure}
 \centering
 \includegraphics[width=8.8cm]{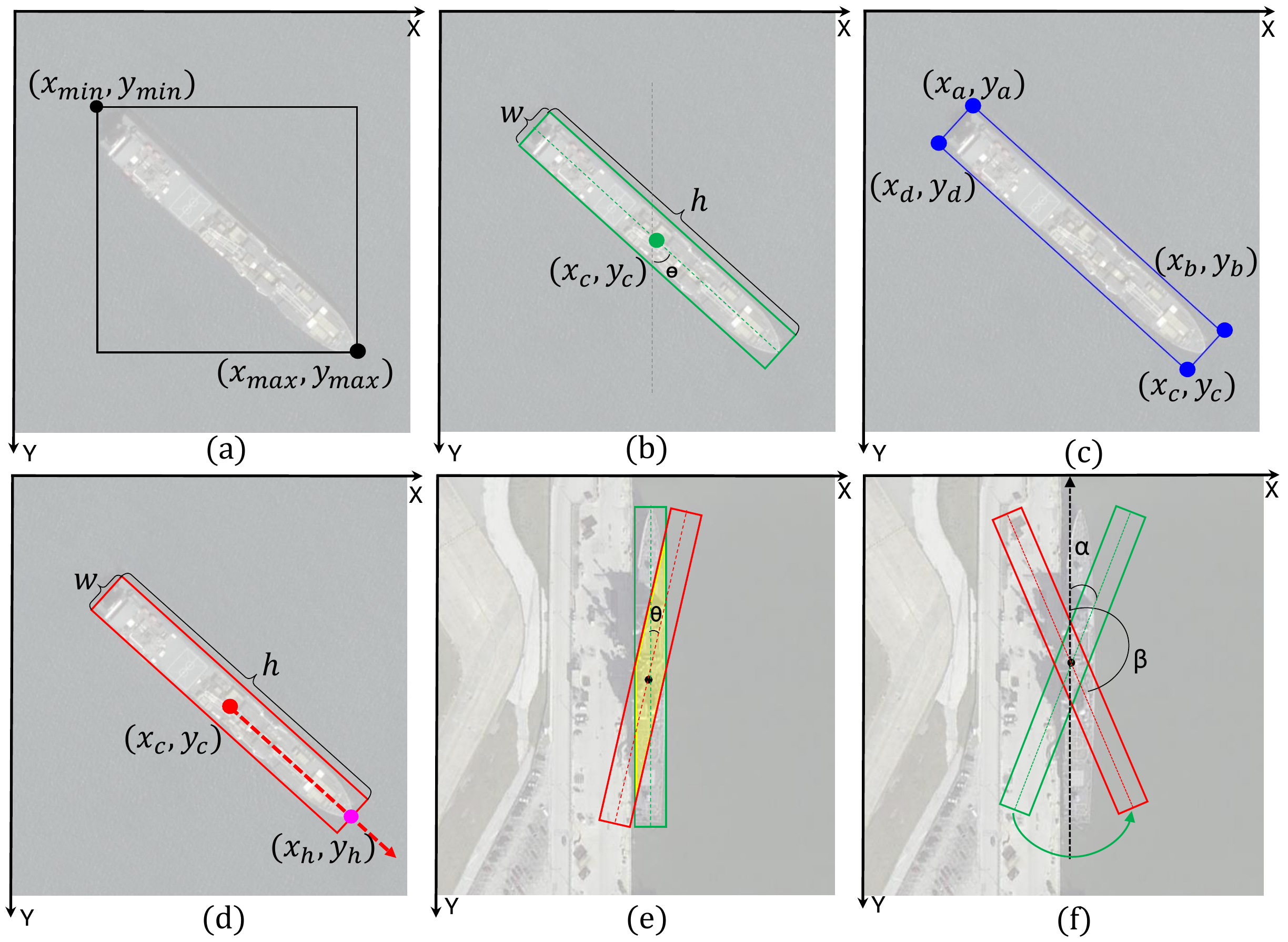}
 \caption{Four different representations of the arbitrary-oriented ship and the disadvantage
 of the angle regression scheme.
 (a) Horizontal boxes parameterized by 4 tuples $(x_{min}, y_{min}, x_{max}, y_{max})$.
 (b) Rotated box with the angle parameterized by 5 tuples $(x_c, y_c, w, h, \theta)$.
 (c) Rotated box with vertices $(a,b,c,d)$, parametrized by 8 tuples $(x_a, y_a, x_b, y_b, x_c, y_c, x_d, y_d)$.
 (d) Rotated box with head point which is parameterized by 6 tuples $(x_c, y_c, w, h, x_h, y_h)$.
 (e) A small angle disturbance will cause a large IoU decrease.
 (f) The angle is discontinous when reaches its range boundary.
 }\label{Reps}
 \end{figure}
 \begin{IEEEkeywords}
 Arbitrary-oriented ship detection, Remote sensing images, Keypoint estimation, Deep convolution neural networks
 \end{IEEEkeywords}

\section{Introduction}\label{introduction}
\IEEEPARstart{S}{hip} detection from high-resolution optical remote sensing images is widely applied in various tasks such as illegal smuggling, port management, and target reconnaissance. Recently, ship detection has received increasing attention and was widely investigated in the past decades \cite{he2021enhancing,li2021gated,deng2019learning, deng2018multi}. However, ship detection in remote sensing images is a highly challenging task due to the arbitrary orientations, densely-parking scenarios, and complex backgrounds \cite{nwpu,sun2021pbnet, he2021dabnet}. To handle the multi-orientation issue, existing methods generally use a series of predefined anchors \cite{li2018rotated}, which has the following shortcomings.

\emph{Inaccurate angle regression.} Fig.~\ref{Reps}(a)-(d) illustrate four different representations of an arbitrary-oriented ship. Since ships in remote sensing images are generally in strips, the intersection over union (IoU) score is very sensitive to the angle of bounding boxes. As shown in Fig.~\ref{Reps}(e), the ground truth box is the bounding box of a ship with an aspect ratio of 10:1.
The red rotated box is generated by rotating the ground truth box with a small angle of $5^{\circ}$. It can be observed that such a small angle variation reduces the IoU between these two boxes to 0.63. Therefore, the anchor-based detectors which define the positive and negative anchors by IoU score usually suffer from an imbalance issue, and thus resulting in detection performance degeneration \cite{qian2019}. Moreover, the angle of the ship is a periodic function, and it is discontinuous at the boundary ($0^{\circ}$ or $180^{\circ}$), as shown in Fig.~\ref{Reps}(f). This discontinuity will also cause performance degeneration.
\begin{figure*}
\centering
\includegraphics[width=17.6cm]{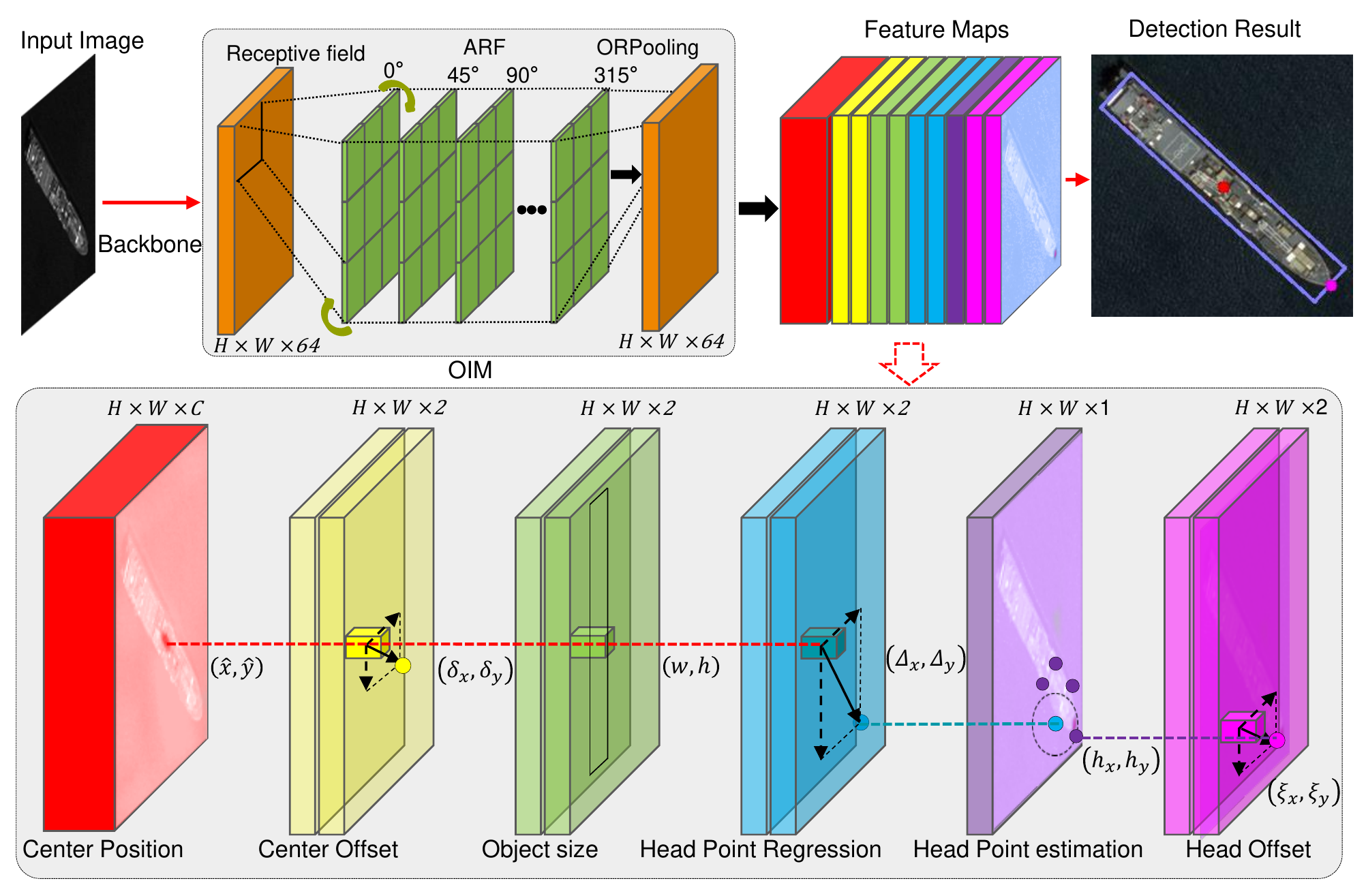}
\caption{The overall framework of our arbitrary-oriented ship detection method. The dotted lines in the graph represent the same position on the feature maps. Feature maps are first generated by using a fully convolutional backbone network and orientation-invariant model (OIM). Afterward, the peaks of the feature map of center points are selected as center points. Then, the center points offset, object sizes, and head regression locations are regressed on the corresponding feature maps at the position of each center point. The potential head points are collected by extracting peaks with confidence scores larger than $0.1$ on the head feature map. The final head location is obtained by assigning each regressed location to its nearest potential head points and then add the head offset.}
\label{framework}
\end{figure*}

\emph{Excessive hyper-parameters and high computational cost.}
Existing methods generally use oriented bounding boxes as anchors to handle rotated objects and thus introduce excessive hyper-parameters such as box sizes, aspect ratios, and orientation angles. Note that, these hyper-parameters have to be manually tuned for novel scenarios, which limits the generalization capability of these methods. Predefined anchor-based methods usually require a large number of anchor boxes. For example, in R$^2$PN \cite{R2PN}, 6 different orientations were used in rotated anchor boxes, and there are a total of 24 anchors at each pixel on its feature maps. A large number of anchor boxes introduce excessive computational cost when calculating IoU scores and executing the non-maximum suppression (NMS) algorithm.

\emph{Under-exploitation of prior information of ships.}

Most previous ship detectors adopted the commonly-used rotation detection algorithms in the area of remote sensing and scene text detection, while overlook the unique characteristics of ships in remote sensing images. That is, the position of the bow is relatively obvious and a certain category of the ship in remote sensing images has a relatively fixed size range by normalizing the ground sample distance (GSD) of images. The size of the ship and the position of the ship's head are important clues for detection. However, these prior informations have been under-exploited by previous ship detection algorithms. These methods only model the ships as rotated rectangles to regress the parameters and do not use the obvious bow point to determine the direction of the ship. Due to the limitation of the effective receptive field of the network, the appearance information near the central point is mainly used in target classification. Size regression and target classification are obtained independently by two parallel branches. Therefore, the size of the target can not effectively assist target classification.

Motivated by the anchor-free detectors CenterNet \cite{Zhou.2019} in natural scenes, in this paper, we propose a one-stage, anchor-free and NMS-free method for arbitrary-oriented ship detection in remote sensing images. We formulate ships as rotated boxes with a head point representing the direction. Specifically, orientation-invariant feature maps are first produced by an orientation-invariant model. Afterward, the peaks of the center feature map are selected as center points. Then, the offset, object sizes, and head positions are regressed on the corresponding feature maps at each center point. Finally, target size is used to adjust the classification score. The architecture of our CHPDet is shown in Fig.~\ref{framework}.

\begin{figure}
 \centering
 \includegraphics[width=8.8cm]{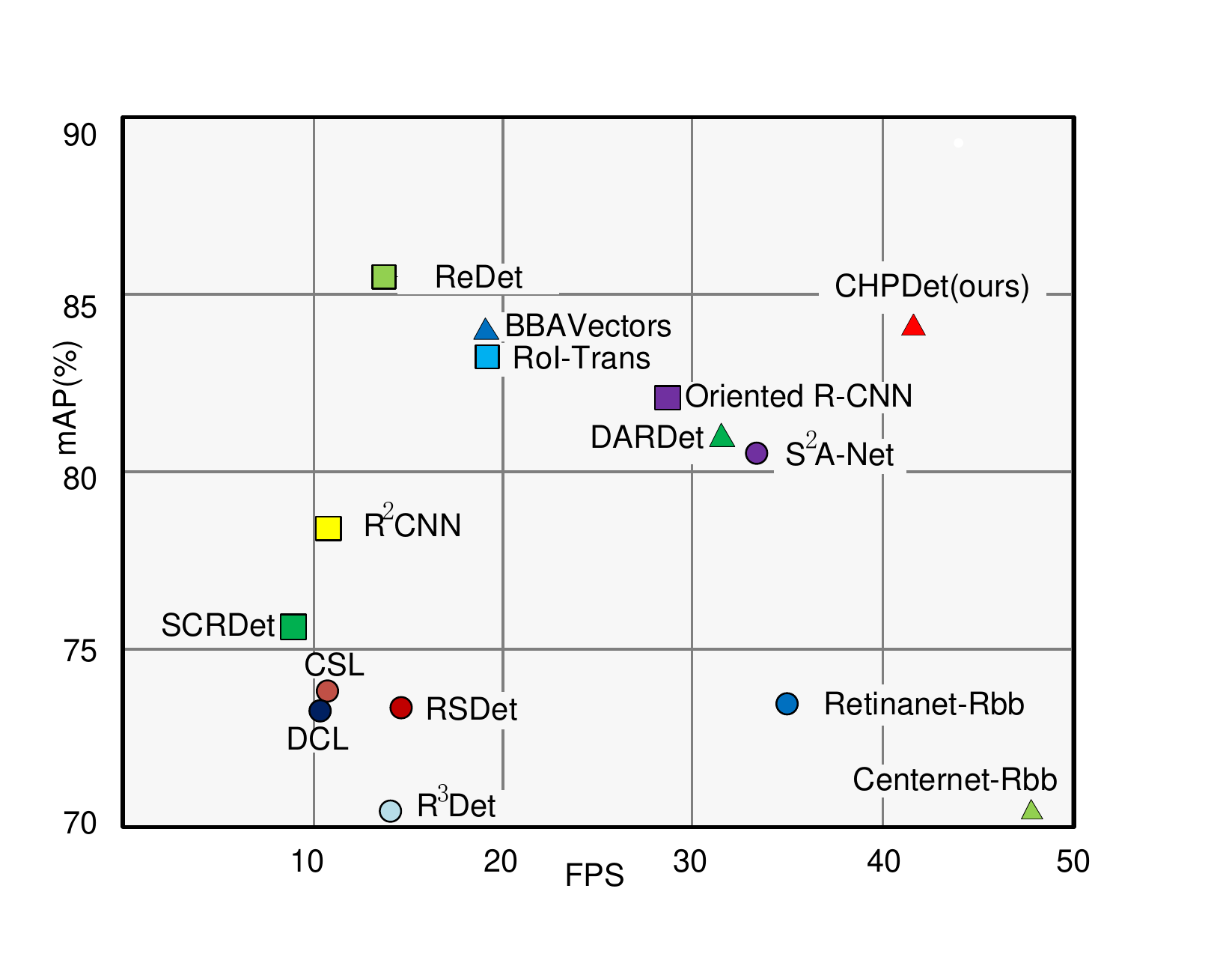}
 \caption{Speed vs. accuracy on our proposed FGSD2021 dataset.}\label{speed}
 \end{figure}

The major contributions of this paper are summarized as follows.
\begin{itemize}

 \item We develop a one-stage, anchor-free ship detector CHPDet, Specifically, we represent the ships using rotated boxes with a head point. This representation addresses the problem of angle periodicity by transforming the angle regression task into a keypoint estimation task. Moreover, our proposed method can expand the scope of angle to [$0^{\circ}$-$360^{\circ}$), and distinguish between bow and stern. 
 \item We design rotated Gaussian kernel to map the annotations into target heatmaps, which can better adapting to the characteristics of the rotated target.

 \item  We propose a module to refine the detection results based on prior information. Moreover, we proposed a new dataset named FGSD2021 for multi-class arbitrary-oriented ship detection in remote sensing images at fixed GSD. This dataset can facilitate the use of prior knowledge of ship size and promote the actual application for remote sensing ship detection.
 \item We introduce an orientation-invariant model (OIM) to generate orientation-invariant feature maps. Extensive experimental results on three datasets show that our CHPDet achieves state-of-the-art performance in both speed and accuracy, as shown in Fig.~\ref{speed}.
\end{itemize}

The rest of this paper is organized as follows. In Section \ref{RelatedWork}, we briefly review the related work. In Section \ref{PROPOSED METHOD}, we introduce the proposed method in detail. Experimental results and analyses are presented in Section \ref{Experiments}. Finally, we conclude this paper in Section \ref{Conclusion}.

\section{Related Work}\label{RelatedWork}
In this section, we briefly review the major works in horizontal object detection, rotated object detection, and remote sensing ship detection.

\subsection{Horizontal Object Detection}
In recent years, deep convolutional neural networks (DCNN) have been developed as a powerful tool for feature representation learning \cite{liu2019unsupervised,maia2021classification}, and have achieved significant improvements in horizontal object detection \cite{liuSurvey}. Existing object detection methods generally represent objects as horizontal boxes, as shown in Fig. \ref{Reps}(a). According to different detection paradigms, deep learning-based object detection methods can be roughly divided into two-stage detectors, single-stage detectors, and multi-stage detectors. Two-stage detectors (e.g., RCNN \cite{girshick2014rich}, Fast-RCNN \cite{girshick2015fast}, Faster-RCNN \cite{ren2015faster}, Mask-RCNN \cite{he2017mask}, R-FCN \cite{dai2016r}) used a pre-processing approach to generate object proposals, and extract features from the generated proposals to predict the category. In contrast, one-stage detectors (e.g., YOLO \cite{redmon2016you,redmon2017yolo9000}, SSD \cite{liu2016ssd}, RetinaNet \cite{lin2017focal}) do not have the pre-processing step and directly performed categorical prediction on the feature maps. Multi-stage detectors (e,g, cascade RCNN \cite{cai2018cascade}, HTC \cite{chen2019hybrid}) performed multiple classifications and regressions, resulting in notable accuracy improvements. In summary, two-stage and multi-stage detectors generally achieve better performance, but one-stage detectors are usually more time-efficient. 

Compared to the above-mentioned anchor-based methods, anchor-free methods \cite{Law.2018} \cite{Zhou.2019} can avoid the requirement of anchors and have become a new research focus in recent years. For example, CornerNet \cite{Law.2018} detected objects at each position of the feature map using the top-left and bottom-right corner points. CenterNet \cite{Zhou.2019} modeled an object as a center point and performed keypoint estimation to find center points and regressed the object size. FCOS \cite{tian2019fcos} predicted four distances, a center score, and a classification score at each position of the feature map to detect objects. The above-mentioned approaches achieve significant improvement in general object detection tasks. However, these detectors can only generate horizontal bounding boxes, which limits their applicability.

\subsection{Arbitrary-oriented object detection}
Arbitrary-oriented detectors are widely used in remote sensing and scene text images. Most of these detectors used rotated bounding boxes or quadrangles to represent multi-oriented objects, as shown in Fig. \ref{Reps}(b) (c). In RRPN \cite{RRPN}, rotated region proposal network was proposed to improve the quality of the region proposals. In R$^2$CNN \cite{r2cnn}, a horizontal region of interest (RoI) was generated to simultaneously predict the horizontal and rotated boxes. RoI-Trans \cite{Ding.2018} transformed a horizontal RoI into a rotated RoI (RRoI). In SCRDet \cite{Yang.2018} and RSDet \cite{qian2019}, novel losses were employed to address the boundary problem for oriented bounding boxes. In R$^3$Det \cite{R3d}, a refined single-stage rotated detector was proposed for the feature misalignment problem. In CSL \cite{CSL} and DCL \cite{DCL}, angle regression was converted into a classification task to handle the boundary problem. In S$^2$A-Net \cite{s2anet}, a fully convolutional layer was proposed to align features to achieve better performance. The aforementioned methods need a set of anchor boxes for classification and regression. These anchors introduce excessive hyper-parameters which limit the generalization capability and introduce an excessive computational cost. At present, several anchor-free arbitrary-oriented detectors (e.g., O$^2$D-Net \cite{middlelines} and X-LineNet \cite{xline}) are proposed to detect oriented objects by predicting a pair of intersecting lines. However, The features used in these methods are not rotation-invariant and the performance still lags behind that of the anchor-base detectors.

\subsection{Ship detection in remote sensing images}
Different from other objects in remote sensing images, ships are in strips with a large aspect ratio. Generally, the outline of the ships is an approximate pentagon with two parallel long sides, and the position of the bow is relatively obvious. Consequently, a certain category of the ship in remote sensing images has a relatively fixed size range by normalizing the GSD of images.

Traditional ship detectors generally used a coarse-to-fine framework with two stages including ship candidate generation and false alarm elimination. For example, Shi et al. \cite{shi2013ship} first generated ship candidates by considering ships as anomalies and then discriminated these candidates using the AdaBoost approach \cite{AdaBoost}. Yang et al. \cite{yang2017ship} proposed a saliency-based method to generate candidate regions, and used a support vector machine (SVM) to further classify these candidates. Liu et al \cite{Liu.2017, Liu.2018} introduced an RRoI pooling layer to extract features of rotated regions. In R$^2$PN \cite{R2PN}, a rotated region proposal network was proposed to generate arbitrary-proposals with ship orientation angle information. The above detectors are also based on a set of anchors and cannot fully exploit the prior information of ships.

\section{Proposed Method}\label{PROPOSED METHOD}

In this section, the architecture of CHPDet is introduced in detail. Our method consists of 5 modules including an arbitrary-oriented ship representation module, a rotated Gaussian kernel module, a head point estimation module, an orientation-invariant module and a probability refinement module. All ships are represented by rotated boxes with a head point. We first detect centers of ships by extracting the peaks in heatmaps which are generated by rotated Gaussian kernels. Then, we locate the head points by two steps (directly regress from image features at the center location, and estimate head points from head heatmaps). We also extract orientation-invariant feature maps by the orientation-invariant model (OIM) to increased consistency between targets and corresponding features. Finally, we refine the detection results based on the prior information. The overall framework of CHPDet is shown in Fig.~\ref{framework}.

\begin{figure}
\centering
\includegraphics[width=4.4cm]{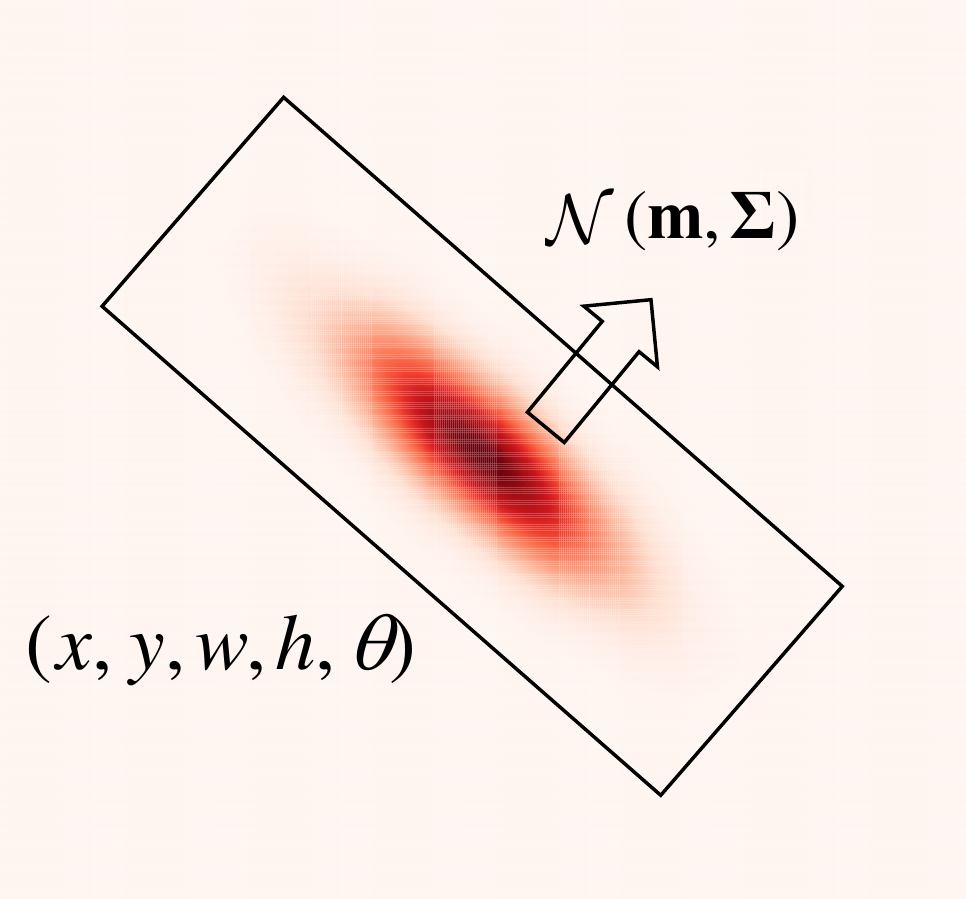}
\caption{A schematic diagram of map a rotated bounding box to a rotated Gaussian distribution.}
\label{guass}
\end{figure}

\begin{figure}
\centering
\includegraphics[width=8.8cm]{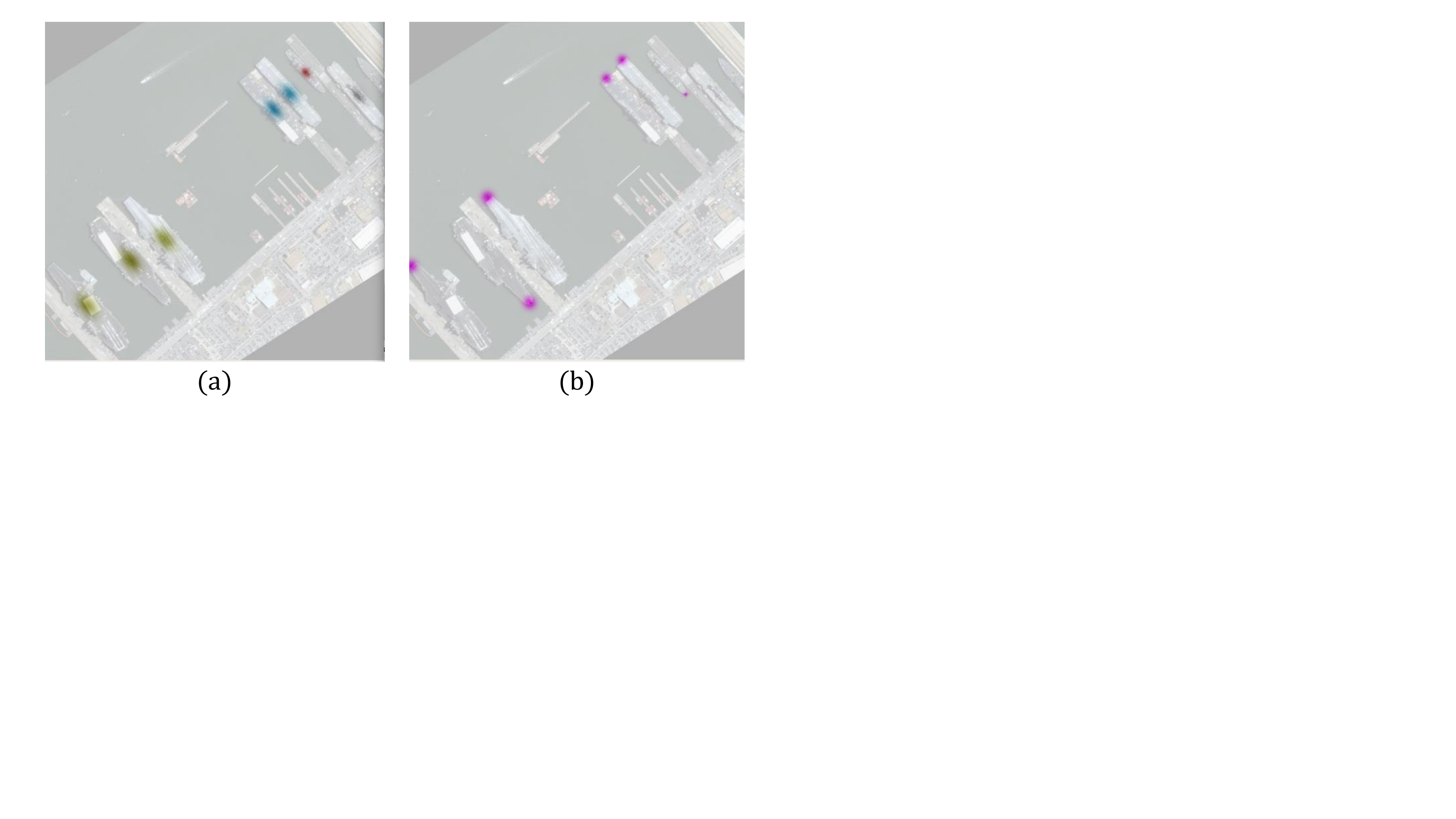}
\caption{A visualization of (a) center heatmap, (b) head heatmap. In center and head heatmaps, different colors represent different categories.}
\label{visualization}
\end{figure}

\subsection{Arbitrary-oriented ship representation}
As shown in Fig.~\ref{Reps}, the widely-used horizontal bounding boxes cannot be directly applied to the arbitrary-oriented ship detection task since excessive redundant background area is included. Moreover, since the arbitrary-oriented ships generally have a large aspect ratio and park densely, the NMS algorithm using a horizontal bounding box tends to produce missing detection. To this end, many methods represent ships as rotated bounding boxes, and these boxes are parameterized with 5 tuples $(c_x, c_y, w, h, \theta)$, where $(x, y)$ is the coordinate of the center of the rotated bounding box, $w$ and $h$ are the width and length of the ship, respectively.
The angle $\theta \in [0^{\circ}, 180^{\circ})$ is the orientation of the long side with respect to the y-axis. This representation can result in the regression inconsistency issue near the boundary case. Recently, some detectors represent objects by four clockwise vertices, which are parameterized by 8 tuples $(x_a, y_a, x_b, y_b, x_c, y_c, x_d, y_d)$. This representation can also introduce regression inconsistency due to the order of the four corner points. 

To avoid the afore-mentioned inconsistency problem, we represent ships as two points and their corresponding size, which are parameterized by 6 tuples $(x_c, y_c, w, h, x_h, y_h)$. $(x_c, y_c)$ is the coordinate of the center of the rotated bounding box, $w$ and $h$ are the width and length of the ship, $(x_h, y_h)$ is the coordinate of the head point of the ship. The direction of the ship is determined by connecting the center and the bow. This representation of ships converts discontinuous angle regression to continuous keypoint estimation. This representation also extends the range of angle representation to $[0^{\circ}, 360^{\circ})$ and enables the network to distinguish between bow and stern.

\subsection{Rotated Gaussian Kernel}\label{CenterDetection}
Our detectors uses center heatmaps to classify and locate ships simultaneously. To adapt to the characteristics of the rotated target, we use the rotated Gaussian kernel (see Fig. \ref{guass}) to map the annotations to target heatmaps in the training stage. 

Specifically, given $m^{th}$ annotated box $\left(x, y,w,h,\theta\right)$  belongs to $c_{m}^{th}$ category, it is linearly mapped to the feature map scale. Then, 2D Gaussian distribution $\mathcal{N}(\mathbf{m}, \mathbf{\Sigma})$ is adopted to produce target heatmap $\textbf{C} \in  \mathbb{R}^{\frac {W} {s}\times \frac {H} {s} \times C}$. Here, $m=(x,y)$ represents the probability density function of the rotated Gaussian distribution, and the probability density function can be calculated according to covariance matrix Eq. \ref{test}.

\begin{equation}\label{test}
\begin{aligned} \Sigma^{1 / 2} &=\mathbf{R S R}^{\top} \\ &=\left(\begin{array}{cc}\cos \theta & -\sin \theta \\ \sin \theta & \cos \theta\end{array}\right)\left(\begin{array}{cc}\sigma _x & 0 \\ 0 & \sigma _y\end{array}\right)\left(\begin{array}{cc}\cos \theta & \sin \theta \\ -\sin \theta & \cos \theta\end{array}\right ) \\ &=\left(\begin{array}{cc}\sigma _x \cos ^{2} \theta+\sigma _y\sin ^{2} \theta & \left(\sigma _x-\sigma _y\right) \cos \theta \sin \theta \\ \left(\sigma _x-\sigma _y\right) \cos \theta \sin \theta & \sigma _x \cos ^{2} \theta+\sigma _y \sin ^{2} \theta\end{array}\right),  \end{aligned}
\end{equation}
where $s$ is a downsampling stride and $\sigma_{x}=\alpha \frac{\sigma_{p}\times w }{\sqrt{w\times h}} $, $ \sigma_{y}=\alpha \frac{\sigma_{p}\times h }{\sqrt{w\times h}} $, $\sigma_{p}$ is a size-adaptive standard deviation \cite{Zhou.2019}. $\alpha$ is set to 1.2 in our implementation, and it’s not carefully selected. Fig.~\ref{guass} is a schematic diagram of mapping a rotated bounding box to a rotated Gaussian distribution.

If two Gaussian kernels belong to the same category with an overlap region, we take the maximum value at each pixel of the feature map. $\hat{\textbf{C}} \in  \mathbb{R}^{\frac {W} {s}\times \frac {H} {s} \times C}$ is a prediction on feature maps produced by the backbones. Fig.~\ref{visualization}(a) shows a visualization of the center heatmaps.

We extract locations with values larger or equal to their 8-connected neighbors as detected center points. The value of the peak point is set as a confidence measurement, and the coordinates in the feature map are used as an index to get other attributes. Therefore, the accurate location of the center point on the feature map is the key part of the whole detection. 

The peaks of the Gaussian kernel, also the centers of rotated box, are treated as the positive samples while any other
pixels are treated as the negative samples, which may cause a huge imbalance between positive and negative samples. To handle the imbalance issue, we use the variant focal loss as \cite{lin2017focal, Zhou.2019}:
\begin{equation}
\mathcal{L}_{c}=\frac{-1}{N}\left\{\begin{array}{cl}\sum_{x y c}\left(1-\hat{\textbf{C}}_{xyc}\right)^{\gamma}\log\left(\hat{\textbf{C}}_{xyc}\right) & \text { if } \textbf{C}(x y c)=1 \\
\sum_{x y c}\left(1-\textbf{C}_{xyc}\right)^{\beta}\left(\hat{\textbf{C}}_{xyc}\right)^{\gamma} \\
\log \left(1-\hat{\textbf{C}}_{xyc}\right) & \text { otherwise }\end{array}\right.
\end{equation}
where $\gamma$ and $\beta$ are the hyper-parameters of the focal loss, $N$ is the number of objects in image $I$ which is used to normalize all positive focal loss instances to $1$. We set $\gamma=2$ and $\beta=4$ in our experiments empirically as in \cite{Law.2018}.

To reduce the quantization error caused by the output stride, we produce local offset feature maps $\textbf{O} \in \mathbb{R}^{\frac{W}{S} \times \frac{H}{S} \times 2}$.
Suppose that $ {c}=\left\{\left(\hat{x}_{k}, \hat{y}_{k}\right)\right\}_{k=1}^{n}$ is the set of detected center points, center point location is given by an integer coordinates $c_k=(\hat{x_i},\hat{y_i})$ on feature map $\textbf{C}$. For each predicted center point $c_k$, let the value on the offset feature maps $f_{k}=(\delta \hat{x}_{k},\delta \hat{y}_{k})$ be the offset of center point $c_k$. The final center point location of class $c$ is $\hat{center_c}=\left\{\left(\hat{x_{k}}+\delta \hat{x}_{k}, \hat{y_{k}}+\delta \hat{y}_{k}\right)\right\}_{k=1}^{n}$. Note that, all classes share the same offset predictions to reduce the computational complexity. The offset is optimized with an L1 loss. This supervision is performed on all center point.
\begin{equation}
\mathcal{L}_{\text {co}}=\frac{1}{N} \sum_{k=1}^{N}\left|\textbf{O}{{c_k}}-\left(\frac{\rm{center}_k}{S}-c_k\right)\right|.
\end{equation}
The regression of the size of objects is similar to that of local offset.

\subsection{Head Point estimation}
We perform two steps for better head points estimation.
\subsubsection{Regression-based head point estimation}
Let $\rm{head}_{k}$$=(h_x,h_y)$ be the $k^{th}$ head point,we directly regress to the offsets $(\varDelta \hat{x}_{k},\varDelta \hat{y}_{k})$ on feature map
$\textbf{R} \in \mathbb{R}^{\frac{W}{S} \times \frac{H}{S} \times 2}$ at each predicted center point $c_k\in \hat{center}$. The regression-based head point is
$\left\{\left(\hat{x_{k}}+\varDelta \hat{x}_{k}, \hat{y_{k}}+\varDelta \hat{y}_{k}\right)\right\}_{k=1}^{n}$, where $\left(\varDelta \hat{x}_{i}, \varDelta \hat{y}_{i}\right)$ is the head point regression, and an L1 loss is used to optimized head regression feature maps.
\begin{equation}
\mathcal{L}_{hr}=\frac{1}{N} \sum_{k=1}^{N}\left|\textbf{R}_{c_k}-h_{k}\right|.
\end{equation}

\subsubsection{Bottom-up head point estimation}
We use standard bottom-up multi-human pose estimation \cite{cao2017realtime} to refine the head points. A target map $\textbf{H} \in  \mathbb{R}^{\frac {W} {s}\times \frac {H} {s} \times 1}$ is computed as described in Section \ref{CenterDetection}. A low-resolution equation is $\rm \tilde{head}=\left\lfloor\frac{head}{s}\right\rfloor$. Head point heatmap $\textbf{E} \in \mathbb{R}^{\frac{W}{S} \times \frac{H}{S} \times 1}$ and local offset heatmap $\textbf{HO} \in \mathbb{R}^{\frac{W}{S} \times \frac{H}{S} \times 2}$ are head maps produced by the backbones. These two head maps are trained with variant focal loss and an L1 loss.
\begin{equation}
\mathcal{L}_{he}=\frac{-1}{N}\sum_{x y}\left\{\begin{array}{cl}\left(1-\textbf{E}_{x y}\right)^{\gamma}\log\left(\textbf{E} _{x y}\right) & \text { if } \textbf{H}_{x y}=1 \\
\left(1-\textbf{H}_{x y}\right)^{\beta}\left(\textbf{E} _{x y}\right)^{\gamma} \\
\log \left(1-\textbf{E} _{x y}\right) & \text { otherwise }\end{array}\right.
\end{equation}
\begin{equation}
\mathcal{L}_{ho}=\frac{1}{N} \sum_{k=1}^{N}\left|\textbf{HO}_{c_k}-\left(\frac{\rm{head_k}}{S}-\tilde{head}\right)\right|.
\end{equation}
The bottom-up head point estimation is the same as the center point detection. Note that, in center point detection, each category has a center points heat map, while in head points estimation, all categories share one head points heatmap. We extract all peak point locations $\hat{\rm{head}}=\left\{\tilde{l}_{i}\right\}_{i=1}$ with a confidence $\textbf{HO}_{x,y} >0.1$ as a potential head points set, and refine the potential head point locations by adding the offset ${(\xi _x,\xi_y)}$. Fig.~\ref{visualization}(b) visualizes the head points heatmap.

We introduce a set of weighted factors to balance the contribution of these parts,
and set $\lambda_{o}=1$, $\lambda_{s}=0.1$, $\lambda_{\rm{hr}}=1$, $\lambda_{\rm{he}}=1$,
and $\lambda_{\rm{ho}}=1$ in all our experiments. We set $\lambda_{s}=0.1$
since the scale of the loss is ranged from $0$ to the output size $ h/S$.
The overall training loss is
\begin{equation}
\begin{aligned}
\mathcal{L}=&\mathcal{L}_{c} +\lambda_{o} \mathcal{L}_{o} +\lambda_{s} \mathcal{L}_{s}+\lambda_{\rm{hr}} \mathcal{L}_{\rm{hr}}+\lambda_{\rm{he}} \mathcal{L}_{\rm{he}}+\lambda_{\rm{ho}} \mathcal{L}_{\rm{ho}}.
\end{aligned}
\end{equation}
In the testing phase, we first extracted the center points on the output center heatmaps $\textbf{C}$ for each category. We used a $3\times3$ max-pooling layer to get the peak points and selected the top 100 peaks as potential center points. Each center point location is represented as an integer coordinates $\hat{c}=(\hat{x},\hat{y})$. Take out the offsets $(\delta_{x},\delta_{y}) $, size $(w,h)$, and head points regression $\left(\varDelta_x, \varDelta_y\right)$ on the corresponding feature map at the location of center points. We also picked all head peak points on the output center heatmaps $\textbf{E}$ with a scores  $\hat{\rm{head}} \in (x,y), if ~\textbf{E}_{x,y} > 0.1$, and then assigned each regressed location ${\rm{head}_{r}=\left(\hat{x}+\varDelta {x}, \hat{y}+\varDelta {y}\right)}$ to its closest detected keypoint $\arg \min _{l \in \rm{head}_{r}}\left(l-\hat{\rm{head}}\right)^{2}$ as the head point $(\hat{h_x},\hat{h_y})$, then we add the head point offset $(\xi_x, \xi _y)$ to refine the head point estimation. Finally, we get the rotated boxes ${(\hat{x}+\delta_x, \hat{y}+\delta_y, w, h, \hat{h_x}+\xi_x,\hat{h_y}+\xi_y )}$. We use the line connecting the center point and the head point to determine the orientation of targets. 
\begin{figure}
\centering
\includegraphics[width=4.4cm]{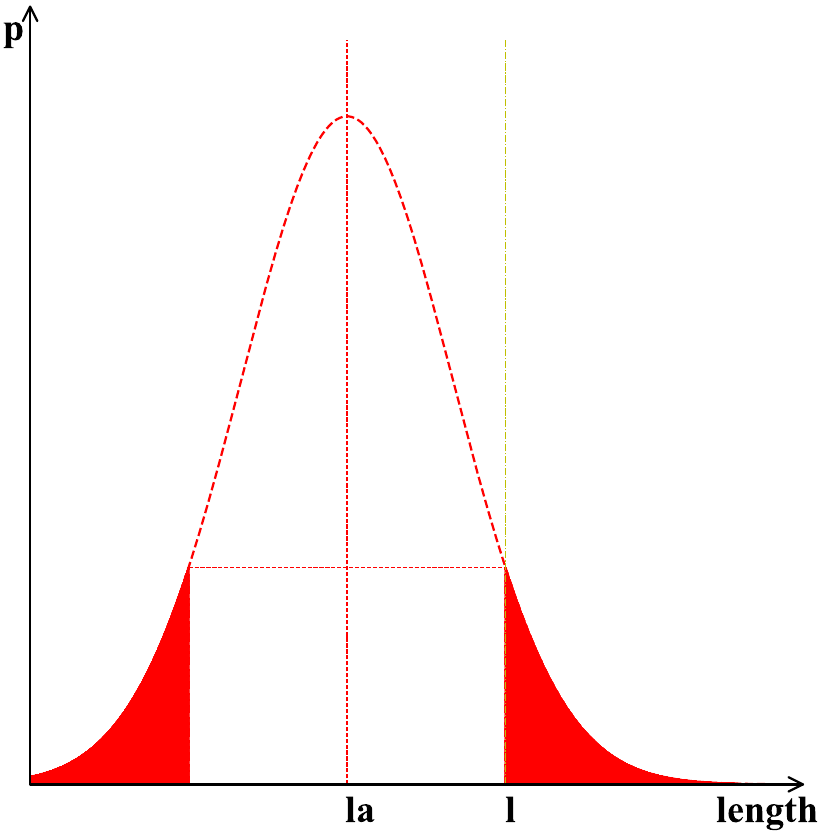}
\caption{A visualization of ship probability density map. In the ship probability density map, $l_a$ represents the mean length of category $a$, $l$ represents the length of the detected ship. The red area is the probability that the target belongs to category $a$.}
\label{refine}
\end{figure}

\subsection{Orientation-Invariant Model}

Let $\textbf{I} \in \mathbb{R}^{W \times H \times 3}$ be an input image with width $W$ and height $H$, the feature map generated from backbone is $\textbf{F} \in \mathbb{R}^{\frac {W} {s}\times \frac {H} {s} \times K}$, where $S$ is the output stride, $C$ is the output feature channels. In this paper, we set the default stride value to $S=4$ and feature channels to $K=64$. 

The feature generated from these backbones is not rotation-invariant \cite{ORN}, while ships in remote sensing images are distributed with arbitrary orientations. To alleviate the inconsistency, we introduce an orientation-invariant model (OIM) which consists of two modules: active rotating filters (ARF) and oriented response pooling (ORPooling) \cite{ORN}. 

We first use active ARF to explicitly encode the orientation information. An ARF is a $k\times k \times N$ filter that actively rotates $N-1$ times during convolution to produce a feature map with $N$ orientation channels. For a feature map \textbf{M} and an ARF $\mathcal{F}$, the $i^{th}$ filter $\mathbf{I}^{(i)}$, $i \in[1, N-1]$, is obtained by clockwise rotating $\mathcal{F}$ by $\frac{2 \pi n}{N}$(N is set to 8 by default) , and can be computed as
\begin{equation}
  \mathbf{I}^{(i)}=\sum_{n=0}^{N-1} \mathcal{F}_{\theta_{i}}^{(n)} \cdot \mathbf{M}^{(n)}, \theta_{i}=i \frac{2 \pi}{N}, i=0, \ldots, N-1
  \end{equation}
  where $\mathcal{F}_{\theta _i }$ is the clockwise $\theta _i$-rotated version of $\mathcal{F}$, $\mathcal{F}_i^{(n)}$
and $\textbf{M}^{(n)}$ are the $n^{th}$ orientation channel of $\mathcal{F}_i$ and $\textbf{M}$ respectively. 
The ARF captures image response in $N$ directions and explicitly encodes its location and orientation into a single feature map with $N$ orientation channels. To reduce computational complexity, we use the combination of small $3 \times 3$ filters and an $8$ orientation channels in our experiments.

Feature maps captured by ARF are not rotation-invariant as orientation information are encoded instead of being discarded. Then ORPooling is used to extract orientation-invariant feature. It is simply achieved by choosing the orientation channel with the strongest response as the output feature $\textbf{I} \in \mathbb{R}^{\frac {W} {s}\times \frac {H} {s} \times K}$. That is, 
\begin{equation}
  \hat{\mathbf{I}}=\max \{\mathbf{I}^{(n)}\}, 0<n<N-1.
\end{equation}
Since ORPooling is introduced to extract the maximum response value for all ARF, the target features of different orientations at this location are identical. Based on the rotation invariance feature, six kinds of feature maps are got by convolution layers respectively. Moreover, OIM only introduces one convolution layer with a small number of parameters, which has little effect on the speed of training and inferencing.

The rotation-invariant feature is very important for detecting arbitrary oriented objects, which enhances the consistency of the feature. Our detectors extract locations with local maximum as detected center points, so at the object center, the rotation-invariant feature of arbitrary oriented objects are identical, which increases the generalization ability of the network. Otherwise, more parameters are needed to encode the orientation information.

\subsection{Refine probability according to size}
By normalizing the GSD of remote sensing images, objects of the same size on the ground have the same size in all images. The size of the target is an important clue to identify the target because a certain type of target in remote sensing images usually has a relatively fixed size range. We propose an approach to adjust the confidence score of targets according to the prior knowledge of ship size. As shown in Fig.~\ref{visualization}(d), suppose that the category of the detected box is $a$, the original confidence score is $s_a$, assume that the length of the detected ship obeys a normal distribution, the mean and standard deviation of the length of category $a$ are $L_a$, $\delta_a$. Then the probability of the target belonging to $a$ is $p_a$, i.e.
\begin{equation}\label{probabilities}
p_a=\frac{2}{\delta_{a} \sqrt{2 \pi}} \int_{-\infty}^{-|l-l a|} \exp \left(-\frac{(x-l a)^{2}}{2 \delta_{a}^{2}}\right) d x.
\end{equation}
To reduce hyper-parameters, we assume that the standard deviation is proportional to the mean $\delta_a=L_a \times \lambda$ for all categories of ships. We multiply the two probabilities to obtain the final detection confidence, $ \hat{p_a}=p_a \times s_a$.

\section{Experiments}\label{Experiments}
We evaluate our method on our FGSD2021 dataset, the public HRSC2016 \cite{HRSC} and UCAS-AOD \cite{aod} dataset. In this section, we first introduce the datasets and implementation details, then perform ablation studies and compare our network to several state-of-the-art methods.

\begin{figure*}
\centering
\includegraphics[width=17.6cm]{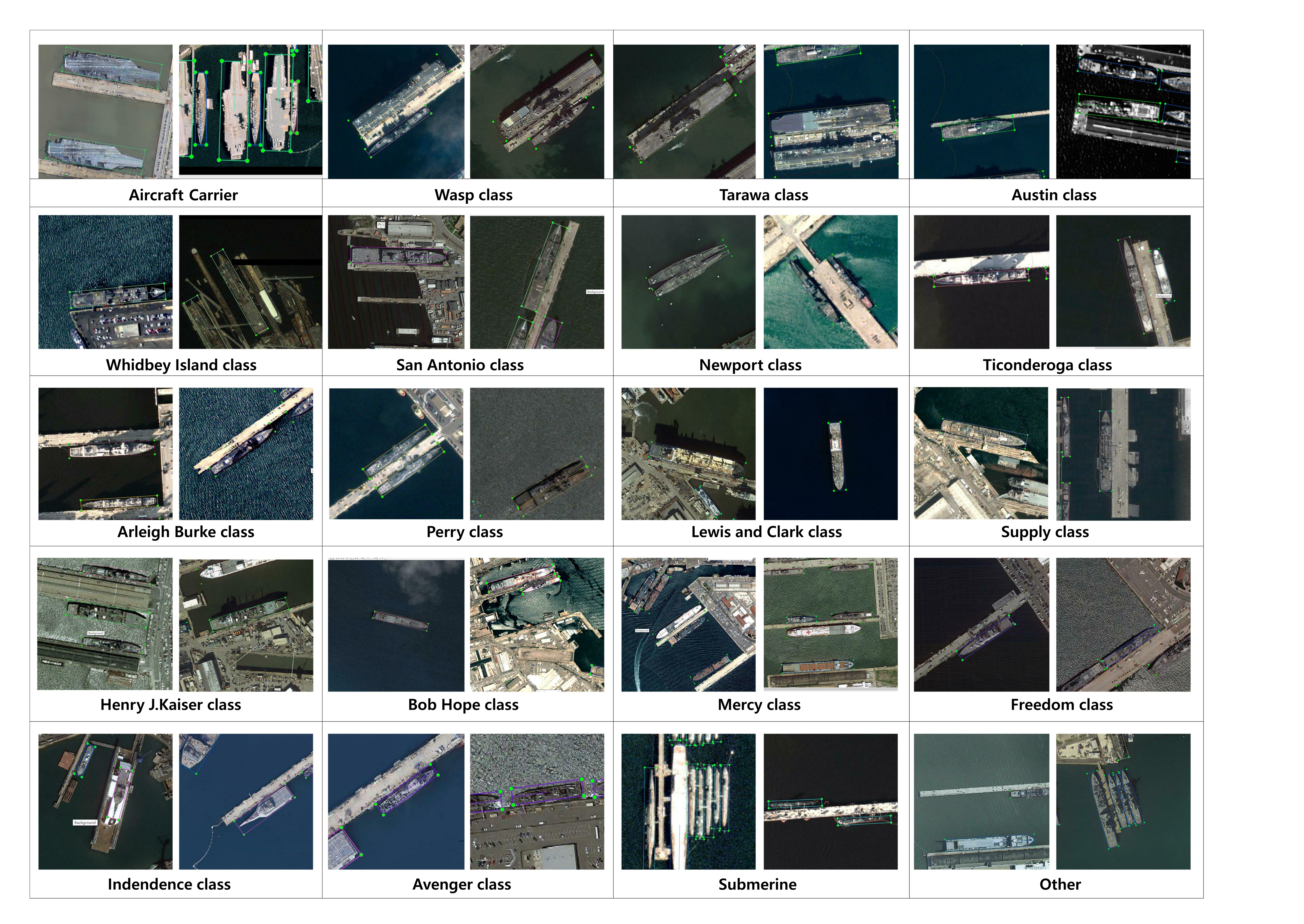}
\caption{Example images from the proposed FGSD2021 dataset. 20 categories are chosen and annotated in our dataset, including Aircraft carriers, Wasp-class, Tarawa-class, Austin-class, Whidbey-island-class, San-Antonio-class, Newport-class, Ticonderoga-class, Arleigh-Burke-class, Perry-class, Lewis and Clark-class, Supply-class, Henry J. Kaiser-class, Bob Hope-Class, Mercy-class, Freedom-class, Independence-class, Avenger-class, submarine, and others.}\label{datasetsample}
\end{figure*}

\subsection{Datasets}
\subsubsection{HRSC2016} The HRSC2016 dataset \cite{HRSC} is a challenging dataset for ship detection in remote sensing images, which collected six famous harbors on Google Earth. The training, validation, and test sets include 436 images with 1207 samples, 181 with 541 samples, and 444 images with 1228 samples, respectively. The image size of this dataset ranges from $300 \times 300$ to $1500 \times 900$. This dataset includes three levels of tasks (i.e., L1, L2, and L3), and these three tasks contain 1 class, 4 classes, and 19 classes, respectively. Besides, the head point of ships is given in this dataset. Following \cite{RRPN} \cite{s2anet} \cite{R3d}, we evaluate our method on task L1. We used the training and validation set in the training phase and evaluated the detection performance on the test set.

\subsubsection{FGSD2021} Existing ship datasets HRSC2016 have the following shortcomings. First, the GSD is unknown, so we cannot get the size of objects in the image by the actual size on the ground. Second, the size of the image is very small which is inconsistent with the actual remote sensing image detection task. To solve these problems, we propose a new ship detection dataset FGSD2021 which has a fixed GSD.

Our dataset is developed by collecting high-resolution satellite images from publicly available Google Earth, which covers some famous ports such as Dandiego, Kitsap-Bremerton, Norfolk, Pearl Harbor, and Yokosuka. We usually obtain multiple images of the same port on different days, and there are also some images from the HRSC2016 dataset. We collected 636 images with a normalized GSD, 1 meter per pixel. The images in our dataset are very large, usually, one image covers a whole port. The width of images is ranged from 157 to 7789 pixels, and the average width is 1202 pixels, the height is ranged from 224 to 6506 pixels, and the average height is 1205 pixels. Our FGSD2021 dataset is divided into 424 training images and 212 test images. The training set is used in the training phase. The detection performance of the proposed method is evaluated on the test set. FGSD2021 including 5274 labeled targets and 20 categories are chosen and annotated. We use the labelimg2\footnote{https://github.com/chinakook/labelImg2} tools to label the ship, the angle range is $[0^{\circ}, 360^{\circ})$, and the main direction is the direction of the bow. Some examples of annotated patches are shown in Fig. \ref{datasetsample}.

\subsubsection{UCAS-AOD} The UCAS-AOD dataset \cite{aod} contains 1510 aerial images of about $659\times1280$ pixels and 14596 instances of two categories including plane and car. The angle range of target in this dataset is $[0^{\circ}, 180^{\circ})$, so we manually marked the direction of the head. We randomly sampled 1132 images for training and 378 images for testing. All images were cropped into patches of size $672 \times 672$.

\subsection{Implementation Details}
Our network was implemented in PyTorch on a PC with Intel Core i7-8700K CPU, NVIDIA RTX 2080Ti GPU. We used the Adam method \cite{kingma2014adam} as the optimizer, and the initial learning rate was set to $2.5 \times 10^{-4}$. We trained our network for 140 epochs with a learning rate being dropped at 90 epochs. During the training phase, we used random rotation, random flipping, and color jittering for data augmentation. To maintain the GSD of the image, we cropped all images into $1024 \times 1024$ slices with a stride of 820,
resized them to $512 \times 512$. We merged the detection results of all the slices to restore the detecting results on the original image. Finally, we applied rotated-non-maximum-suppression (RNMS) with an IoU threshold of 0.15 to discard repetitive detections. The speed of the proposed network was measured on a single NVIDIA RTX 2080Ti GPU.

Several different backbones (e.g., deep layer aggregation (DLA) \cite{2017Deep} and hourglass network (Hourglass) \cite{2016Stacked}) can be used to extract features from images. We followed CenterNet \cite{Zhou.2019} to enhance DLA by replacing ordinary convolutions with deformable convolutions and add a 256 channel $3 \times 3$ convolutional layer before the output head. The hourglass network consists of two sequential hourglass modules. Each hourglass module includes 5 pairs of down and up convolutional networks with skip connections. This network generally yields better keypoint estimation performance \cite{Law.2018}.

\subsection{Evaluation Metrics}
The IoU between oriented boxes is used to distinguish detection results. The mean average precision (mAP) and head direction accuracy are used to evaluate the performance of arbitrary-Oriented detectors.
\subsubsection{IoU} The IoU is the result of dividing the overlapping area by the union area of two boxes. We adopted the evaluation approach in DOTA \cite{dota} to get the IoU. If the IoU between a detection box and a ground-truth is higher than a threshold, the detection box is marked as true-positive (TP), otherwise false-positive (FP). If a ground-truth box has no matching detections, it is marked as false negative (FN).
\subsubsection{mAP} The precision and recall are calculate by $\text {precision }=\frac{\text { TP }}{\mathrm{TP}+\mathrm{FP}}$, $\text {recall}=\frac{\text {TP}}{\mathrm{TP}+\mathrm{FN}}$. We first set a set of thresholds, and then we get a corresponding maximum precision for each recall threshold. AP is the average of these precisions. The mean average precision (mAP) is the mean of APs over all classes. The mAP$_{0.5}$-mAP$_{0.8}$ is computed under the IoU threshold of 0.5-0.8 respectively. PASCAL VOC2007 metric is used to compute the mAP in all of our experiments.
\subsubsection{Head direction accuracy} The prediction angle range of the previous algorithm is 0$^{\circ}$-180$^{\circ}$, which cannot distinguish between the bow and stern of the ship. The mAP base on the IoU between two rotated boxes is taken as the only evaluation criterion, which cannot reflect the accuracy of the bow direction. To solve this problem, we define bow direction accuracy as an additional evaluation. That is the proportion of the ships whose angle difference from the ground-truth less than 10 degrees in all TPs.
\subsection{Ablation Study}
In this subsection, we present ablation experiments to investigate our models.

\subsubsection{CenterNet as baseline} As an anchor-free detector, CenterNet performs keypoint estimation to find the center point and regresses the object size at each center point position. To carry out arbitrary-oriented ship detection, we add an extra branch to predict the angle as a baseline which is named CenterNet-Rbb. CenterNet-Rbb uses a DLA34 as the backbone, and presents ships as rotated boxes with angle, and uses the L1 loss function to optimized angle regression feature maps. We set weighted factor $\lambda_{angle}=0.1$ to balance the contribution of these parts since the scale of the loss is ranged from $0$ to $180$. As shown in Table \ref{Refine probability}, CenterNet-Rbb achieves an mAP of 70.52\% which demonstrates that our baseline achieves competitive performance.

\begin{table}
  \caption{Results achieved on FGSD2021 with different ablation versions.
  `\textit{Baseline}' represents adding a branch to predict the angle based on CenterNet.   
  `\textit{Head Point}' represents replacing the angle prediction branch to head point estimation module.
  `\textit{Rotate kernel}' represents generating center heatmap by rotated kernel in training.
  `\textit{OIM}' represents add orientation-invariant model behand the backbone.
  `\textit{Extra convolution}' represents replacing the OIM with two extra convolution layers.
  `\textit{Refine probability}' represents using the prior size information to adjust the confidence score of the detected boxes.}\label{Refine probability}
  \begin{tabular}{p{2.1cm}<{\centering}|p{0.7cm}<{\centering}|ccccc}
  \Xhline{1pt}
                   & baseline  & \multicolumn{4}{c}{Different Settings of CHPDet} \\
  \Xhline{1pt}
  Head Point        &           & $\checkmark$ &$\checkmark$&$\checkmark$&$\checkmark$&$\checkmark$\\
  Rotate kernel     &           &              &$\checkmark$&$\checkmark$&$\checkmark$&$\checkmark$\\
  OIM               &           &              &            &$\checkmark$&            &$\checkmark$\\
  Extra convolution &           &              &            &            &$\checkmark$&            \\
  Refine Probability&           &              &            &            &            &$\checkmark$\\
  \hline
  mAP               &  70.52    & 82.96        & 83.56      & 86.61      & 82.66      & 87.91    \\    
  \Xhline{1pt}
  \end{tabular}
  \end{table}

  \begin{table*}
  \centering
  \caption{Performance of CHEDet achieved on FGSD2021 with different variance coefficient $\lambda$.
  `\textit{without refine}' represents using the original confidence without refinement.
  `\textit{Ground truth class}' represents using ground truth class label to eliminate
  the misclassification.}\label{coefficient}
  \centering
  \begin{tabular}{c|c|c|c|c|c|c|c|c|c|c|c}
  \Xhline{1pt}
    \multirow{2}{*}{Backbone}& \multirow{2}{*}{Image Size} &\multicolumn{8}{c|}{coefficient $\lambda$}& \multirow{2}{*}{without refine}& \multirow{2}{*}{Ground truth class} \\
  \cline{3-10}
                              &    & 0.1   &  0.2  & 0.3   & 0.4   & 0.5   & 0.6   & 0.7   & 0.8   &   &   \\
  \Xhline{1pt}
  DLA34        & $512 \times 512$  & 87.40 & \textbf{87.91} & 87.39 & 87.45 & 87.17 & 87.20 & 87.15 & 87.10  &86.61 & 89.33 \\
  DLA34      & $1024 \times 1024$  & 86.37 & 87.84 & \textbf{89.28} & 88.17 & 88.68 & 88.85 & 88.47 & 88.50  & 88.39 & 89.74 \\
  \Xhline{1pt}
  \end{tabular}
  \end{table*}
  
\begin{table*}
\centering
\caption{Detection accuracy on different types of ships and overall performance with the state-of-the-art methods on FGSD. The short names for categories are defined as (abbreviation-full name): Air - Aircraft carriers, Was - Wasp class, Tar - Tarawa class, Aus - Austin class, Whi - Whidbey Island class, San -San Antonio class, New - Newport class, Tic - Ticonderoga class, Bur- Arleigh Burke class, Per - Perry class, Lew -Lewis and Clark class, Sup - Supply class, Kai - Henry J. Kaiser class, Hop - Bob Hope Class, Mer - Mercy class, Fre - Freedom class, Ind - Independence class, Ave - Avenger class, Sub - Submarine and Oth - Other. CHPDet$^{\dagger}$ means CHPDet trained and detected with $1024 \times 1024$ image size.} \label{proposed accuracy}
\begin{tabular}{c|p{0.26cm} p{0.26cm} p{0.26cm} p{0.26cm} p{0.26cm} p{0.26cm} p{0.26cm} p{0.26cm} p{0.26cm} p{0.26cm} p{0.26cm} p{0.26cm} p{0.26cm} p{0.26cm} p{0.26cm} p{0.26cm} p{0.26cm} p{0.26cm} p{0.26cm} p{0.4cm} |c <{\centering}}
\Xhline{1pt}
Method                       & Air & Was & Tar& Aus & Whi & San & New & Tic & Bur & Per & Lew & Sup & Kai & Hop & Mer & Fre & Ind & Ave & Sub  & Oth  & mAP(07) \\
\Xhline{1pt}
R$^{2}$CNN \cite{r2cnn}  & 89.9& 80.9& 80.5& 79.4& 87.0& 87.8& 44.2 & 89.0& 89.6&79.5& \textcolor{red}{80.4}& 47.7& 81.5& 87.4& \textcolor{red}{100} & 82.4& \textcolor{red}{100} & 66.4 &50.9& 57.2&  78.09 \\
Retinanet-Rbb \cite{lin2017focal} & 89.7& 89.2& 78.2& 87.3& 77.0& 86.9& 62.7& 81.5& 83.3& 70.6& 46.8& 69.9& 80.2& 83.1& \textcolor{red}{100} & 80.6& 89.7& 61.5& 42.5&  9.1& 73.49 \\
ROI-Trans\cite{Ding.2018}    & \textcolor{blue}{90.9}& 88.6& 87.2& 89.5& 78.5& 88.8& 81.8& 89.6& 89.8& \textcolor{blue}{90.4}& 71.7& 74.7& 73.7& 81.6& 78.6& \textcolor{red}{100} & 75.6& 78.4& 68.0& \textcolor{blue}{66.9}& 83.48  \\
SCRDet  \cite{Yang.2018}     & 77.3& 90.4& 87.4& 89.8& 78.8& \textcolor{blue}{90.9}& 54.5& 88.3& 89.6& 74.9& 68.4& 59.2& 90.4& 77.2& 81.8& 73.9& \textcolor{red}{100} & 43.9& 43.8& 57.1&75.90  \\
CSL \cite{CSL}              & 89.7& 81.3& 77.2& 80.2& 71.4& 77.2& 52.7& 87.7& 87.7& 74.2& 57.1& \textcolor{red}{97.2}& 77.6& 80.5& \textcolor{red}{100} & 72.7& \textcolor{red}{100} & 32.6& 37.0& 40.7& 73.73  \\
DCL  \cite{DCL}              & 89.9& 81.4& 78.6& 80.7& 78.0& 87.9& 49.8& 78.7& 87.2& 76.1& 60.6& 76.9& \textcolor{blue}{90.4}& 80.0& 78.8& 77.9& \textcolor{red}{100} & 37.1& 31.2& 45.6& 73.34  \\
R$^3$Det\cite{R3d}           & \textcolor{blue}{90.9}& 80.9& 81.5& 90.1& 79.3& 87.5& 29.5& 77.4& 89.4& 69.7& 59.9& 67.3& 80.7& 76.8& 72.7& 83.3& \textcolor{blue}{90.9}& 38.4& 23.1& 40.0& 70.47  \\
RSDet\cite{qian2019}         & 89.8& 80.4& 75.8& 77.3& 78.6& 88.8& 26.1& 84.7& 87.6& 75.2& 55.1& 74.4& 89.7& 89.3& \textcolor{red}{100} &86.4 & \textcolor{red}{100} & 27.6& 37.6& 50.6& 73.74  \\
S$^{2}$A-Net\cite{s2anet}    & \textcolor{blue}{90.9}& 81.4& 73.3& 89.1& 80.9& 89.9& 81.2& 89.2& \textcolor{red}{90.7}& 88.9& 60.5& 75.9& 81.6& 89.2& \textcolor{red}{100} & 68.6& \textcolor{blue}{90.9}& 61.3& 55.7& 64.7& 80.19  \\
ReDet\cite{redet}            & \textcolor{blue}{90.9}& \textcolor{red}{90.6}& 80.3& 81.5& \textcolor{blue}{89.3}& 88.4& 81.8& 88.8& 90.3& \textcolor{red}{90.5}& 78.1& 76.0& \textcolor{red}{90.7}& 87.0& \textcolor{blue}{98.2}& 84.4& \textcolor{blue}{90.9}& 74.6& \textcolor{blue}{85.3}& 71.2& 85.44  \\
Oriented R-CNN\cite{Oriented}& \textcolor{blue}{90.9}& 89.7& 81.5& 81.1& 79.6& 88.2& \textcolor{blue}{98.9}& 89.8& \textcolor{blue}{90.6}& 87.8& 60.4& 73.9& 81.8& 86.7& \textcolor{red}{100} & 60.0 &\textcolor{red}{100} & 79.4& 66.9& 63.7& 82.54  \\
BBAVectors\cite{BBAVectors}  & \textcolor{red}{99.5}& \textcolor{red}{90.9}& 75.9& \textcolor{red}{94.3}& 90.9& 52.9& 88.5& \textcolor{blue}{90.0}& 80.4& 72.2& \textcolor{blue}{76.9}& 88.2& 99.6& \textcolor{red}{100}& 94.0 & 100 & 74.5& 58.9& 63.1& \textcolor{red}{81.1}& 83.59  \\
DARDet\cite{DARDet}          & \textcolor{blue}{90.9}& 89.2& 69.7& 89.6& 88.0& 81.4& 90.3& 89.5& 90.5& 79.7& 62.5& 87.9& 90.2& 89.2& \textcolor{red}{100} & 68.9& 81.8& 66.3& 44.3& 56.2& 80.31  \\

\hline
CenterNet-Rbb\cite{Zhou.2019} & 67.2& 77.9& 79.2& 75.5& 66.8& 79.8& 76.8& 83.1& 89.0& 77.7& 54.5& 72.6& 77.4& \textcolor{red}{100} & \textcolor{red}{100}& 60.8& 74.8& 46.5& 44.1& 6.8 & 70.52 \\
CHPDet-DLA34                  & \textcolor{blue}{90.9}& 90.4& \textcolor{blue}{89.6}& 89.3& \textcolor{red}{89.6}& \textcolor{red}{99.1}& \textcolor{red}{99.4}& \textcolor{red}{90.2}& 90.2& 90.3& 70.7& 87.9& 89.2& \textcolor{blue}{96.5}& \textcolor{red}{100} & 85.1& \textcolor{red}{100}& \textcolor{blue}{84.4}& 68.5& 56.9& \textcolor{blue}{87.91}  \\
CHPDet-DLA34$^{\dagger}$      & \textcolor{blue}{90.9}& 90.2& \textcolor{red}{90.9}& \textcolor{blue}{90.3}& \textcolor{blue}{89.3}& 89.2& \textcolor{blue}{98.9}& \textcolor{red}{90.2}& 90.2& 90.2& 72.2&\textcolor{blue}{96.5}& \textcolor{red}{90.7}& 95.3& \textcolor{red}{100} & \textcolor{blue}{95.2}& \textcolor{blue}{90.9}& \textcolor{red}{86.4}& \textcolor{red}{85.9}& 62.4& \textcolor{red}{89.29}  \\
\Xhline{1pt}
\end{tabular}
\end{table*}

\subsubsection{Effectiveness of the head point estimation} When we replace the angle prediction branch with the head point estimation module, the overall mAP is improved from 70.52\% to 82.96\%. It is a significant improvement, which fully demonstrates the effectiveness of the head point estimation approach. This improvement mainly comes from two aspects. First, the algorithm makes full use of the prior knowledge of the bow point and improves the accuracy of angle regression. Second, since multi-task learning is performed, bow detection increases the supervision information and improves the accuracy of other tasks. 

To further verify the promoting effect of head point estimation for center point detection and size detection, we set all angles of ground-truth and the detected box to 0$^{\circ}$. Compared with the CenterNet-Rbb, The mAP of CHPDet has risen from 84.4\% to 88.0\%. This shows that the head point estimation is equivalent to multi-task joint training. It gives more supervision to the network and improves the performance of the network. Besides, the head point estimation only introduces 3 additional channels feature maps and 0.7 ms speed latency. 

\subsubsection{Effectiveness of the rotated Gaussian kernel} Our detector uses the rotated Gaussian kernel to map the annotations to target heatmaps and achieves an improvement of 0.6\% in terms of nomal Gaussian kernels. This implies that rotated Gaussian kernel is a better representation for OBB in the aerial images. 

The rotated Gaussian kernel can adjust its shape and direction according to the shape of the target and reduce the influence of positioning error on the detection results. As shown in Fig.~\ref{guass}, the rotated Gaussian kernel has the maximum error in the long axis direction, so in the detection process, the center point has a large error on the long axis. Because the error of the center point in the long axis has the least influence on the IoU, the rotated Gaussian kernel can reduce the influence of positioning error on the detection results, and vice versa. Note that, rotated Gaussian kernel does not introduce any additional parameters, and they do not increase training and inferencing time. Consequently, it is a completely cost-free module.

\subsubsection{Effectiveness of the orientation-invariant model} We add an orientation-invariant model (OIM) at the end of the backbone and keep other settings unchanged to validate its effectiveness. As shown in Table \ref{Refine probability}, compared with the standard backbone, the backbone with the orientation-invariant model improves mAP by about 3 percentages to 86.61\%, while only introduces 2.6 ms speed latency. 

To further verify the effectiveness of the OIM structure, we replace the OIM with two convolution layers. Compared with the standard backbone, the backbone with two extra convolution layers model drops the performance to 82.66\%. It is proved that the performance improvement does not come from the improvement of the number of parameters. 

We argue that the standard backbones are not rotation-invariant, and the corresponding features are rotation-sensitive. Consequently, OIM increases the consistency between targets and corresponding features. It not only improves the accuracy of angle prediction, but also improves the accuracy of center point detection and size regression.

\subsubsection{Effectiveness of the Refine probability model}  In the FGSD2021 dataset, the actual length of each category is determined. For example, the length of the Ticonderoga-class cruiser is 172.8 meters. In our designed network,  the prior knowledge of ship length is used to refine the confidence of the detected ships belonging to a certain category. Table \ref{Refine probability} shows the mAP values of different ablation versions on the test set. It can be observed that the baseline model achieves the lowest mAP. When the prior size information is incorporated, the performance has been improved. The accuracy improvement on low-resolution images is more obvious, e.g., from 86.61\% to 87.91\%, an increase of 1.3\% in mAP. It demonstrates that the prior size information can improve classification accuracy.

We set a variance coefficient to adjust the influence of size on probability. Consequently, we use the length of this type of ship $l_a$ multiplied by a coefficient $r$ as the mean square error of this type $\delta_{a}$, $\delta_{a}=l_a \times r$. The variance coefficient will affect classification accuracy. When the coefficient is large, the probability difference between different categories will be smaller, and the influence of the size on the confidence of the category will be smaller, and vice versa. As can be observed in Table \ref{coefficient}, when the coefficient is small, it is equivalent to use size as the main information to classify objects. Accuracy increases gradually as the coefficient increases, and when the coefficient is larger than 0.2, the coefficient has little impact on the accuracy. When we treat all categories as one category and remove the category influence on the detection results, the mAP is $89.33$\%, and $89.74$\%, respectively. At the same time, by incorporating prior information to adjust the classification confidence, the detection accuracy under 20 categories with an input image of size 1024x1024 achieved an mAP of $89.28$\% which shows that after incorporating the prior information, almost all categories are classified correctly.

\begin{table*}
\centering
\caption{Detection performance on the FGSD2021 at different IoU thresholds and the accuracy of bow direction. BDA presents bow direction accuracy}\label{proposed result}
\begin{tabular}{c |c |c|c|c|c|c|c|  p{2cm}<{\centering}}
\Xhline{1pt}
Method     & Backbone  & Image Size &mAP$_{0.5}$&mAP$_{0.6}$&mAP$_{0.7}$&mAP$_{0.8}$& BDA& FPS   \\
\Xhline{1pt}
R$^{2}$CNN\cite{r2cnn}        & Resnet50  & $512 \times 512$      & 78.09 & 75.03& 64.83& 36.41&\_& 10.3 \\
Retinanet-Rbb\cite{lin2017focal}& Resnet50& $512 \times 512$      & 73.49 & 69.17& 62.82& 45.00&\_& 35.6 \\
RoI-Trans\cite{Ding.2018}     & Resnet50  & $512 \times 512$      & 83.48 & 82.63& 80.35& 65.18&\_& 19.2 \\
SCRDet\cite{Yang.2018}        & Resnet50  & $512 \times 512$      & 75.90 & 70.98& 61.82& 35.12&\_ & 9.2 \\
CSL\cite{CSL}                 & Resnet50  & $512 \times 512$       & 73.73 & 69.71& 60.25& 34.93&\_& 10.4\\
DCL\cite{DCL}                 & Resnet50  & $512 \times 512$     & 73.34 & 69.19& 57.80& 28.54&\_ & 10.0 \\
R$^3$Det\cite{R3d}            & Resnet50  & $512 \times 512$       & 70.47 & 68.32& 57.17& 27.44&\_& 14.0\\
RSDet\cite{qian2019}          & Resnet50  & $512 \times 512$       & 73.74 & 69.55& 61.52& 35.83&\_& 15.4\\
S$^2$A-Net\cite{s2anet}       & Resnet50  & $512 \times 512$      & 80.19 & 79.58& 75.65& 58.82&\_ & 33.1\\
ReDet\cite{redet}             & ReResnet50& $512 \times 512$      & 85.44 & 84.65& 80.24& 67.94&\_& 13.8 \\
Oriented R-CNN\cite{Oriented} & Resnet50  & $512 \times 512$      & 82.54 & 81.32& 78.53& 64.87&\_& 27.4 \\
BBAVectors\cite{BBAVectors}   & Resnet50  & $512 \times 512$       & 83.59 & 82.74& 78.55& 62.48&\_& 18.5\\
DARDet\cite{DARDet}           & Resnet50  & $512 \times 512$     & 80.31 & 79.62& 74.77& 59.21&\_ & 31.9 \\
\hline
CenterNet-Rbb\cite{Zhou.2019} & DLA34     & $512 \times 512$   & 70.52 & 69.34& 65.52& 45.33&\_ & \textcolor{red}{48.5}\\
CHPDet(ours)                  & DLA34     & $512 \times 512$   & \textcolor{blue}{87.91} & \textcolor{blue}{87.15}& \textcolor{blue}{83.69}& \textcolor{blue}{71.24}&\textcolor{blue}{97.84} & \textcolor{blue}{41.7}\\
CHPDet(ours)                  & DLA34     & $1024 \times 1024$    &\textcolor{red}{89.29} & \textcolor{red}{88.98}& \textcolor{red}{86.57}& \textcolor{red}{73.56}&\textcolor{red}{98.39}& 15.4 \\
\Xhline{1pt}
\end{tabular}
\end{table*}

\begin{figure*}
\centering
\includegraphics[width=17.6cm]{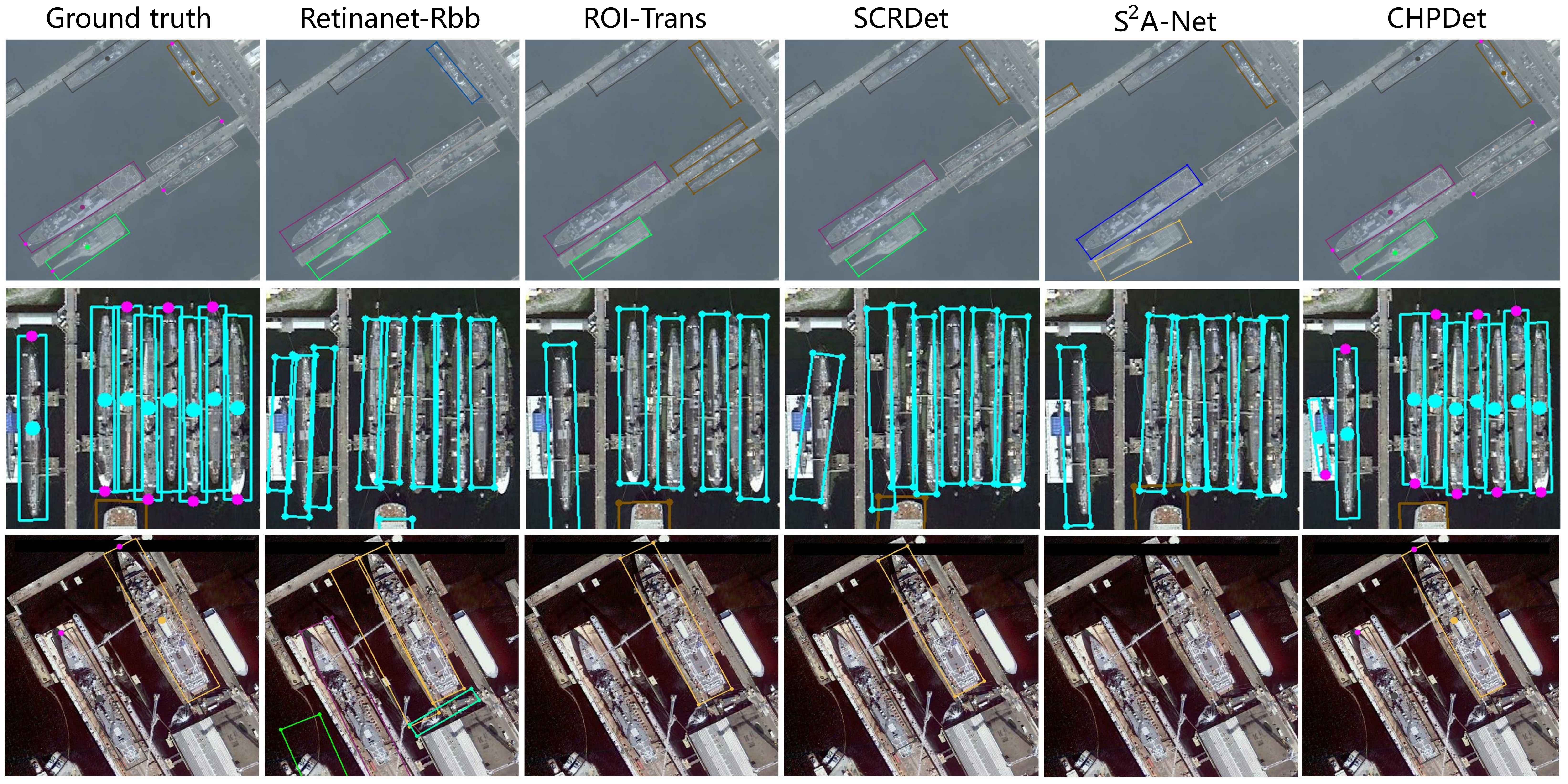}
\caption{Comparison of the detection results in FGSD2021 with different methods. The first column is the ground truth, and the second to the last columns are the results of Retinanet-Rbb \cite{lin2017focal}, ROI-Trans \cite{Ding.2018}, SCRDet \cite{Yang.2018} , S$^2$A-Net \cite{s2anet}, and CHPDet (ours), respectively. Different color of rotated boxes represents a different type of ships. The pink point represents the head point.}\label{example result}
\end{figure*}

\begin{table}
  \centering
  \caption{Detection accuracy on the HRSC2016 dataset, 07 means using the 2007 evaluation metric.}\label{HRSC-result}
  \begin{tabular}{c|c|c}
  \Xhline{1pt}
  Method                         & Backbone          & mAP(07) \\
  \Xhline{1pt}
  R$^{2}$CNN \cite{r2cnn}        & Resnet101       & 73.07 \\
  RRPN \cite{RRPN}               & Resnet101       & 79.08 \\
  R$^{2}$PN\cite{R2PN}           & VGG16           & 79.6 \\
  ROI-trans\cite{Ding.2018}      & Resnet101       & 86.20 \\
  Gliding Vertex\cite{Gliding}   & Resnet101       & 88.20 \\
  BBAVectors\cite{BBAVectors}    & Resnet101       & 88.6 \\
  R$^3$Det \cite{R3d}            & Resnet101       & 89.26 \\
  FPN-CSL\cite{CSL}              & Resnet101       & 89.62\\
  R$^3$Det-DCL\cite{DCL}         & Resnet101       & 89.46\\
  DAL\cite{DAL}                  & Resnet101       & 89.77 \\
  R$^3$Det-GWD \cite{GWD}        & Resnet101       & 89.85 \\
  RSDet \cite{qian2019}          & ResNet152       & 86.5 \\
  FR-Est \cite{pointbased}       & Resnet101       & 89.7 \\
  S$^{2}$A-Net \cite{s2anet}     & Resnet101       & 90.2 \\
  Oriented RepPoints\cite{OrientedRepPoints}& Resnet50  & 90.38 \\
  ReDet\cite{redet}              & ReResnet50      & 90.46 \\
  Oriented R-CNN\cite{Oriented}  & Resnet101       & \textcolor{blue}{90.50} \\
  DARDet \cite{DARDet}            & Resnet50       & 90.37 \\
  \hline
  CHPDet(ours)                    & DLA34          & 88.81 \\
  CHPDet(ours)                    & Hourglass104  &\textcolor{red}{90.55}\\
  \Xhline{1pt}
  \end{tabular}
  \end{table}

\begin{table}
  \centering
  \caption{Detection accuracy on the UCAS-AOD dataset.}\label{AOD-result}
  \begin{tabular}{c|c|c|c|c}
  \Xhline{1pt}
  Method                         & Backbone  & car  & airplane &mAP(07) \\
  \Xhline{1pt}
  YOLOv3 \cite{yolov3}           & Darknet53 &74.63  &89.52    &82.08       \\
  RetinaNet \cite{lin2017focal}  & Resnet101 &84.64  &90.51    &87.57        \\
  FR-O\cite{dota}                & Resnet101 &86.87  &89.86    &88.36        \\
  ROI-trans\cite{Ding.2018}      & Resnet101 &87.99  &89.90    &88.95        \\
  FPN-CSL\cite{CSL}              & Resnet101 &88.09  &90.38    &89.23         \\
  R$^3$Det-DCL\cite{DCL}         & Resnet101 &88.15  &90.57    &89.36       \\
  DAL\cite{DAL}                  & Resnet101 &\textcolor{red}{89.25}  &90.49    &\textcolor{blue}{89.87}        \\
  \hline
  CHPDet(ours)                   & DLA34      &88.58  &\textcolor{blue}{90.64}    &89.61        \\
  CHPDet(ours)                   & Hourglass104 &\textcolor{blue}{89.18}  &\textcolor{red}{90.81}  &\textcolor{red}{90.00}        \\
  \Xhline{1pt}
  \end{tabular}
  \end{table}

\subsubsection{Bow direction accuracy} 
\begin{figure}
  \centering
  \includegraphics[width=8.8cm]{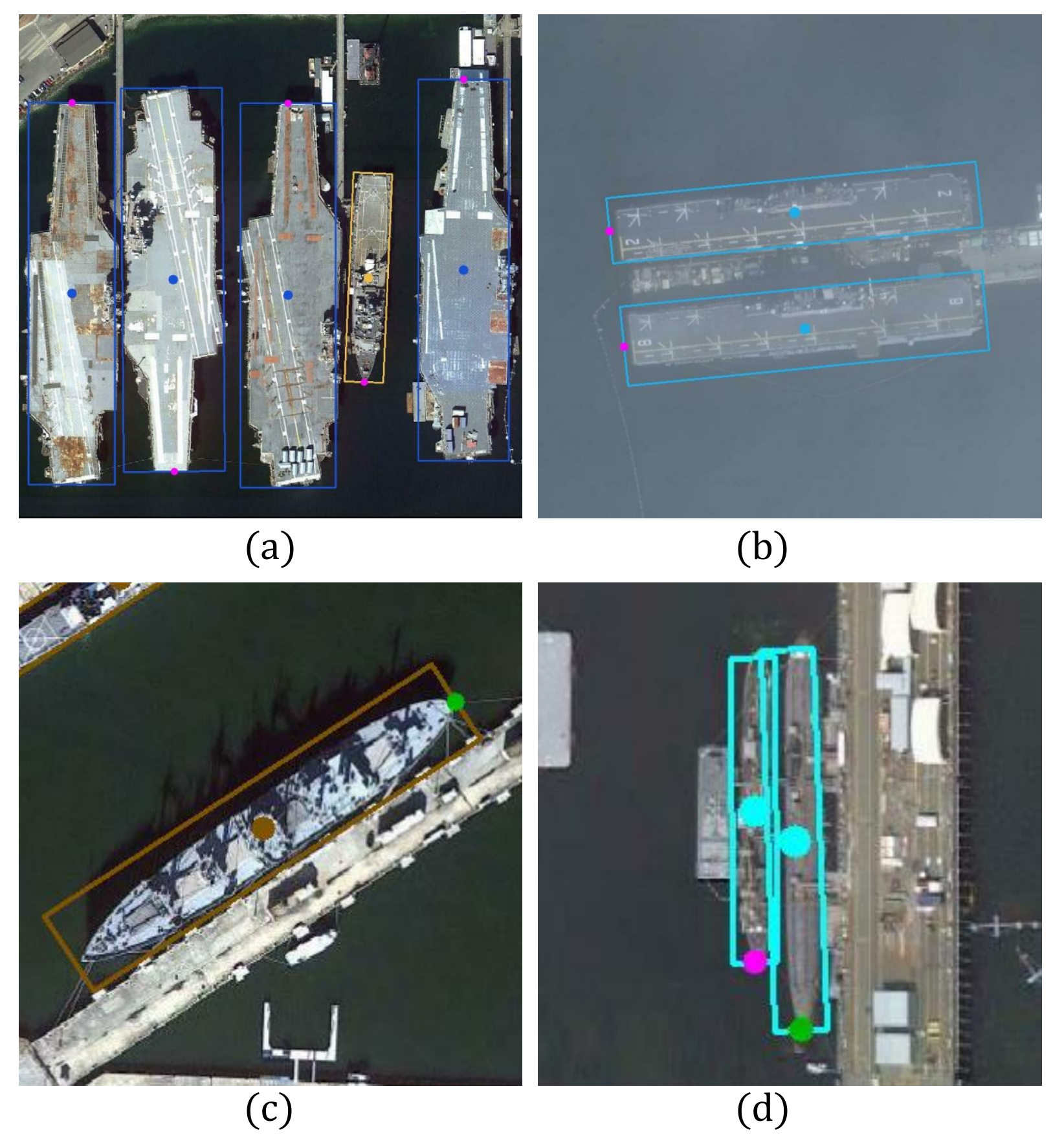}
  \caption{Some bow direction detection result of CHPDet. The pink dots represent the correct head point and the green dots represent the wrong head point.}\label{head_detect}
  \end{figure}

It can be seen from Table \ref{proposed accuracy} that the bow direction accuracy of our CHPDet is up to 97.84, 98.14, and 98.39, respectively. This shows that almost all bow directions of ships are correct. As shown in Fig.~\ref{head_detect}, the pink dots represent the correct head point and the green dots represent the wrong head point. Our detection algorithm can well detect the bow direction of all types of ships, including aircraft carriers, amphibious ships. Only a small number of ships or submarines whose bow and stern are similar from a bird-view perspective, the bow direction will be opposite.

\subsection{Comparison with other methods}
In this section, we compare our method with other representative ship detectors including RetinaNet-Rbb \cite{lin2017focal} ROI-trans \cite{Ding.2018}\footnote{https://github.com/dingjiansw101/AerialDetection/}, R$^2$CNN \cite{r2cnn}, CSL \cite{CSL}, DCL \cite{DCL}, RSDet \cite{qian2019}, SCRDet \cite{Yang.2018}\footnote{https://github.com/yangxue0827/RotationDetection}, and S$^2$A-Net \cite{s2anet}\footnote{https://github.com/csuhan/s2anet} on three benchmark datasets including FGSD2021, HRSC2016 \cite{HRSC} and UCAS-AOD \cite{aod}. To achieve fair comparison, we used the default settings of the original codes on the DOTA dataset including the same data augmentation strategy, and the number of training epochs.

\subsubsection{Results on FGSD2021}
We evaluate CHPDet on the FGSD2021 dataset and compare our method with other rotation detection methods. 
It can be seen from Table \ref{proposed accuracy} that CHPDet achieves $87.91\%$ mAP at the speed of $41.7$ FPS, which surpass the other compared methods. 
Compared with the general rotation detection methods RoI-Trans \cite{Ding.2018} and S$^2$A-Net \cite{s2anet}, our proposed method achieves a remarkable improvement
by 4.5\%, 7.7\% in mAP and 19.3, 8.6 in FPS. When higher resolution images are used, the accuracy can be improved to $89.29\%$. This confirms that our method achieves a large superiority in terms of accuracy and speed. To further verify the accuracy of the prediction, we gradually increase the IoU threshold. As can be seen from Table \ref{proposed result}, when the IoU threshold is gradually increased, the performance of other detectors dropped significantly, and the decline of our detector is relatively small. When the IoU threshold was increased to $0.8$, the mAP of our CHPDet remained at $71.24$. This shows that our detector can get higher quality rotated boxes than other algorithms.

Fig.~\ref{example result} shows a visual comparison of the detection results of Retinanet-Rbb \cite{lin2017focal}, ROI-Trans \cite{Ding.2018}, SCRDet \cite{Yang.2018}, S$^2$A-Net \cite{s2anet}, and our method. As shown in the first row, all the other methods have misclassification or false alarms, S$^2$A-Net \cite{s2anet} has an inaccurate angle prediction, while our method precisely detects them. For the densely parking scene in the second row, all the compared detectors lost at least two submarines, and our method is not influenced by the densely parking scene. The last row of Fig.~\ref{example result} is a harbor with a complex background. Note that, two ships are not in the water but on the dry dock. ROI-trans \cite{Ding.2018} and S$^2$A-Net \cite{s2anet} miss the targets, SCRDet \cite{Yang.2018} has an inaccurate bounding box. Compared to these four methods, our method can better detect the ships in the complex background and is more robust for challenging situations. 

This improvement mainly comes from three aspects. First, the algorithm makes full use of the prior knowledge of the bow point and improves the accuracy of directional regression. Second, since multi-task learning is performed, bow detection increases the supervision information and improves the accuracy of other tasks. Last, the prior knowledge of ship length is used to refine the confidence of the detected ships belonging to a certain category.  The usage of the prior knowledge of ships introduces significant performance improvements.

\begin{figure*}
\centering
\includegraphics[width=17.6cm]{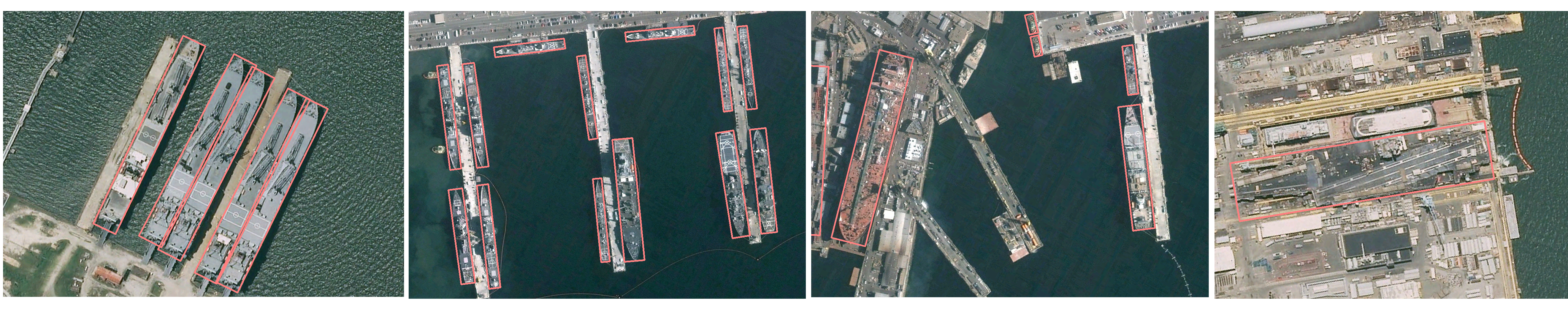}
\caption{Sample object detection results of our proposed CHPDet on HRSC2016 dataset.}\label{result_hrsc}
\end{figure*}

\begin{figure*}
\centering
\includegraphics[width=17.6cm]{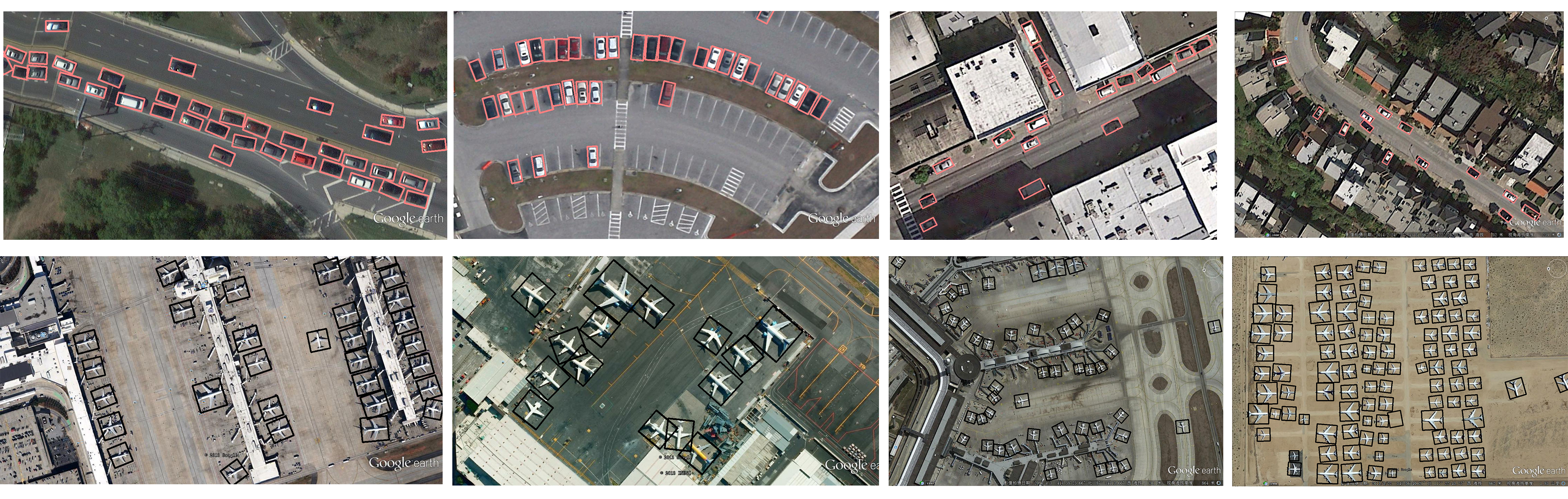}
\caption{Sample object detection results of our proposed CHPDet on UCAS-AOD dataset.}\label{result_aod}
\end{figure*}

\subsubsection{Results on HRSC2016}

The HRSC2016 dataset contains plenty of ships with arbitrary orientations. we evaluate our method on task L1 which contains 1 class and report the results with VOC2007 metric. 
To demonstrate the performance of our detector, we compare it with other state-of-the-art methods, i.e., ReDet \cite{redet}, Oriented R-CNN \cite{Oriented}, and Oriented RepPoints \cite{OrientedRepPoints}.
The overall comparison performance is reported in Table \ref{HRSC-result}. Our method achieves the best performance over all the compared methods, at an accuracy of $90.55\%$.
To further show the performance of CHPDet, the detection results are visualized in Fig. \ref{result_hrsc}. As shown in the first two columns, the densely parked ships can be detected well. In the last two columns, there is a lot of background around ships, which is a huge challenge for detectors. The results indicate that our proposed method can avoid false alarms in complex background.

\subsubsection{Results on UCAS-AOD} 

The UCAS-AOD dataset contains a large mount of cars and planes, which are often overwhelmed by a complex background in aerial images.
For a fair comparison, we only report the results under VOC2007 metric. Table \ref{AOD-result} shows the results with the recent methods on the UCAS-AOD dataset. It can be seen that our proposed method achieves the best performance (with an mAP of 90.00\%). The CHPDet, which uses a larger output resolution (output stride of 4) compared to traditional object detectors (output stride of 8) and presents ship as the center and head points, can capture abundant information of small objects. Fig. \ref{result_aod} gives some example detection results on the UCAS-AOD dataset. We find that CHPDet performs well in a variety of challenging scenes, which demonstrates the generalization capability of the detector.

\section{Conclusion}\label{Conclusion}

Our proposed approach converts discontinuous angle regression to continuous keypoint estimation by formulating ships as rotated boxes with a head point representing the direction. This design can incorporate the prior knowledge of the bow point, which not only improves the detection performance, but also expands the scope of predicted angle to $[0^{\circ}-360^{\circ})$. Our method can distinguish between bow and stern.
CHPDet has simple structure. It has only one positive sample per annotation and simply extracts local peaks in the keypoint heatmap. It does not need Non-Maximum Suppression (NMS). This design ensures high time efficiency.
The prior knowledge of ship length is also incorporated to refine the confidence of the detected ships belonging to a certain category.

Although our method achieves encouraging results on ship detection from remote sensing images, our method can not be directly used in normal object detection datasets in aerial images
such as DOTA \cite{dota}. That is because, CHPDet needs more accurate annotations which mark the direction of the target head in the range of $360^{\circ}$.
CHPDet is several times faster than most detectors in inference, but it suffers from a long training time. For future work, we will address this issue by encoding more training samples from annotated boxes.

In this paper, we proposed a one-stage anchor-free detection framework to detect arbitrary-oriented ships from remote sensing images by making full use of the prior of ships. Our method detects ships by extracting the center, head of ships, and regresses the size of ships at each center point with rotation-invariant features. And we refine the detection results based on the prior information.  And we refine the detection results based on the prior information. CHPDet avoids complex anchor design and computing relative to the anchor-based methods and can accurately predict angles in a large range [$0^{\circ}$-$360^{\circ}$). Experimental results demonstrate that our method achieves better accuracy and efficiency as compared with other ship detectors.

\bibliographystyle{IEEEtran}
\bibliography{TGRS}

\begin{thebibliography}{10}
\providecommand{\url}[1]{#1}
\csname url@samestyle\endcsname
\providecommand{\newblock}{\relax}
\providecommand{\bibinfo}[2]{#2}
\providecommand{\BIBentrySTDinterwordspacing}{\spaceskip=0pt\relax}
\providecommand{\BIBentryALTinterwordstretchfactor}{4}
\providecommand{\BIBentryALTinterwordspacing}{\spaceskip=\fontdimen2\font plus
\BIBentryALTinterwordstretchfactor\fontdimen3\font minus
  \fontdimen4\font\relax}
\providecommand{\BIBforeignlanguage}[2]{{%
\expandafter\ifx\csname l@#1\endcsname\relax
\typeout{** WARNING: IEEEtran.bst: No hyphenation pattern has been}%
\typeout{** loaded for the language `#1'. Using the pattern for}%
\typeout{** the default language instead.}%
\else
\language=\csname l@#1\endcsname
\fi
#2}}
\providecommand{\BIBdecl}{\relax}
\BIBdecl

\bibitem{he2021enhancing}
S.~He, H.~Zou, Y.~Wang, R.~Li, F.~Cheng, X.~Cao, and M.~Li, ``Enhancing
  mid–low-resolution ship detection with high-resolution feature
  distillation,'' \emph{IEEE Geoscience and Remote Sensing Letters}, 2021.

\bibitem{li2021gated}
B.~Li, Y.~Guo, J.~Yang, L.~Wang, Y.~Wang, and W.~An, ``Gated recurrent
  multiattention network for vhr remote sensing image classification,''
  \emph{IEEE Transactions on Geoscience and Remote Sensing}, 2021.

\bibitem{deng2019learning}
Z.~Deng, H.~Sun, S.~Zhou, and J.~Zhao, ``Learning deep ship detector in sar
  images from scratch,'' \emph{IEEE Transactions on Geoscience and Remote
  Sensing}, vol.~57, no.~6, pp. 4021--4039, 2019.

\bibitem{deng2018multi}
Z.~Deng, H.~Sun, S.~Zhou, J.~Zhao, L.~Lei, and H.~Zou, ``Multi-scale object
  detection in remote sensing imagery with convolutional neural networks,''
  \emph{ISPRS Journal of Photogrammetry and Remote Sensing}, vol. 145, pp.
  3--22, 2018.

\bibitem{nwpu}
G.~Cheng and J.~Han, ``A survey on object detection in optical remote sensing
  images,'' \emph{ISPRS Journal of Photogrammetry and Remote Sensing}, vol.
  117, pp. 11--28, 2016.

\bibitem{sun2021pbnet}
X.~Sun, P.~Wang, C.~Wang, Y.~Liu, and K.~Fu, ``Pbnet: Part-based convolutional
  neural network for complex composite object detection in remote sensing
  imagery,'' \emph{ISPRS Journal of Photogrammetry and Remote Sensing}, vol.
  173, pp. 50--65, 2021.

\bibitem{he2021dabnet}
Q.~He, X.~Sun, Z.~Yan, and K.~Fu, ``Dabnet: Deformable contextual and
  boundary-weighted network for cloud detection in remote sensing images,''
  \emph{IEEE Transactions on Geoscience and Remote Sensing}, 2021.

\bibitem{li2018rotated}
M.~Li, W.~Guo, Z.~Zhang, W.~Yu, and T.~Zhang, ``Rotated region based fully
  convolutional network for ship detection,'' in \emph{IGARSS 2018-2018 IEEE
  International Geoscience and Remote Sensing Symposium}.\hskip 1em plus 0.5em
  minus 0.4em\relax IEEE, 2018, pp. 673--676.

\bibitem{qian2019}
W.~Qian, X.~Yang, S.~Peng, Y.~Guo, and C.~Yan, ``Learning modulated loss for
  rotated object detection,'' \emph{arXiv preprint arXiv:1911.08299}, 2019.

\bibitem{R2PN}
Z.~Zhang, W.~Guo, S.~Zhu, and W.~Yu, ``Toward arbitrary-oriented ship detection
  with rotated region proposal and discrimination networks,'' \emph{IEEE
  Geoscience and Remote Sensing Letters}, vol.~15, no.~11, pp. 1745--1749,
  2018.

\bibitem{Zhou.2019}
X.~Zhou, D.~Wang, and P.~Kr{\"a}henb{\"u}hl, ``Objects as points,'' \emph{arXiv
  preprint arXiv:1904.07850}, 2019.

\bibitem{liu2019unsupervised}
S.~Liu, Q.~Du, X.~Tong, A.~Samat, and L.~Bruzzone, ``Unsupervised change
  detection in multispectral remote sensing images via spectral-spatial band
  expansion,'' \emph{IEEE Journal of Selected Topics in Applied Earth
  Observations and Remote Sensing}, vol.~12, no.~9, pp. 3578--3587, 2019.

\bibitem{maia2021classification}
D.~S. Maia, M.-T. Pham, E.~Aptoula, F.~Guiotte, and S.~Lef{\`e}vre,
  ``Classification of remote sensing data with morphological attributes
  profiles: a decade of advances,'' \emph{IEEE Geoscience and Remote Sensing
  Magazine}, 2021.

\bibitem{liuSurvey}
L.~Liu, W.~Ouyang, X.~Wang, P.~Fieguth, J.~Chen, X.~Liu, and
  M.~Pietik{\"a}inen, ``Deep learning for generic object detection: A survey,''
  \emph{International Journal of Computer Vision}, vol. 128, no.~2, pp.
  261--318, 2020.

\bibitem{girshick2014rich}
R.~B. Girshick, J.~Donahue, T.~Darrell, and J.~Malik, ``Rich feature
  hierarchies for accurate object detection and semantic segmentation,'' in
  \emph{2014 {IEEE} Conference on Computer Vision and Pattern Recognition,
  {CVPR} 2014, Columbus, OH, USA, June 23-28, 2014}.\hskip 1em plus 0.5em minus
  0.4em\relax {IEEE} Computer Society, 2014, pp. 580--587.

\bibitem{girshick2015fast}
R.~B. Girshick, ``Fast {R-CNN},'' in \emph{2015 {IEEE} International Conference
  on Computer Vision, {ICCV} 2015, Santiago, Chile, December 7-13, 2015}.\hskip
  1em plus 0.5em minus 0.4em\relax {IEEE} Computer Society, 2015, pp.
  1440--1448.

\bibitem{ren2015faster}
S.~Ren, K.~He, R.~B. Girshick, and J.~Sun, ``Faster {R-CNN:} towards real-time
  object detection with region proposal networks,'' in \emph{Advances in Neural
  Information Processing Systems 28: Annual Conference on Neural Information
  Processing Systems 2015, December 7-12, 2015, Montreal, Quebec, Canada},
  C.~Cortes, N.~D. Lawrence, D.~D. Lee, M.~Sugiyama, and R.~Garnett, Eds.,
  2015, pp. 91--99.

\bibitem{he2017mask}
K.~He, G.~Gkioxari, P.~Doll{\'{a}}r, and R.~B. Girshick, ``Mask {R-CNN},'' in
  \emph{{IEEE} International Conference on Computer Vision, {ICCV} 2017,
  Venice, Italy, October 22-29, 2017}.\hskip 1em plus 0.5em minus 0.4em\relax
  {IEEE} Computer Society, 2017, pp. 2980--2988.

\bibitem{dai2016r}
J.~Dai, Y.~Li, K.~He, and J.~Sun, ``{R-FCN:} object detection via region-based
  fully convolutional networks,'' in \emph{Advances in Neural Information
  Processing Systems 29: Annual Conference on Neural Information Processing
  Systems 2016, December 5-10, 2016, Barcelona, Spain}, D.~D. Lee, M.~Sugiyama,
  U.~von Luxburg, I.~Guyon, and R.~Garnett, Eds., 2016, pp. 379--387.

\bibitem{redmon2016you}
J.~Redmon, S.~K. Divvala, R.~B. Girshick, and A.~Farhadi, ``You only look once:
  Unified, real-time object detection,'' in \emph{2016 {IEEE} Conference on
  Computer Vision and Pattern Recognition, {CVPR} 2016, Las Vegas, NV, USA,
  June 27-30, 2016}.\hskip 1em plus 0.5em minus 0.4em\relax {IEEE} Computer
  Society, 2016, pp. 779--788.

\bibitem{redmon2017yolo9000}
J.~Redmon and A.~Farhadi, ``{YOLO9000:} better, faster, stronger,'' in
  \emph{2017 {IEEE} Conference on Computer Vision and Pattern Recognition,
  {CVPR} 2017, Honolulu, HI, USA, July 21-26, 2017}.\hskip 1em plus 0.5em minus
  0.4em\relax {IEEE} Computer Society, 2017, pp. 6517--6525.

\bibitem{liu2016ssd}
W.~Liu, D.~Anguelov, D.~Erhan, C.~Szegedy, S.~Reed, C.-Y. Fu, and A.~C. Berg,
  ``Ssd: Single shot multibox detector,'' in \emph{European Conference on
  Computer Vision}.\hskip 1em plus 0.5em minus 0.4em\relax Springer, 2016, pp.
  21--37.

\bibitem{lin2017focal}
T.~Lin, P.~Goyal, R.~B. Girshick, K.~He, and P.~Doll{\'{a}}r, ``Focal loss for
  dense object detection,'' in \emph{{IEEE} International Conference on
  Computer Vision, {ICCV} 2017, Venice, Italy, October 22-29, 2017}.\hskip 1em
  plus 0.5em minus 0.4em\relax {IEEE} Computer Society, 2017, pp. 2999--3007.

\bibitem{cai2018cascade}
Z.~Cai and N.~Vasconcelos, ``Cascade {R-CNN:} delving into high quality object
  detection,'' in \emph{2018 {IEEE} Conference on Computer Vision and Pattern
  Recognition, {CVPR} 2018, Salt Lake City, UT, USA, June 18-22, 2018}.\hskip
  1em plus 0.5em minus 0.4em\relax {IEEE} Computer Society, 2018, pp.
  6154--6162.

\bibitem{chen2019hybrid}
K.~Chen, J.~Pang, J.~Wang, Y.~Xiong, X.~Li, S.~Sun, W.~Feng, Z.~Liu, J.~Shi,
  W.~Ouyang, C.~C. Loy, and D.~Lin, ``Hybrid task cascade for instance
  segmentation,'' in \emph{{IEEE} Conference on Computer Vision and Pattern
  Recognition, {CVPR} 2019, Long Beach, CA, USA, June 16-20, 2019}.\hskip 1em
  plus 0.5em minus 0.4em\relax Computer Vision Foundation / {IEEE}, 2019, pp.
  4974--4983.

\bibitem{Law.2018}
H.~Law and J.~Deng, ``Cornernet: Detecting objects as paired keypoints,'' in
  \emph{Proceedings of the European Conference on Computer Vision (ECCV)},
  2018, pp. 734--750.

\bibitem{tian2019fcos}
Z.~Tian, C.~Shen, H.~Chen, and T.~He, ``{FCOS:} fully convolutional one-stage
  object detection,'' in \emph{2019 {IEEE/CVF} International Conference on
  Computer Vision, {ICCV} 2019, Seoul, Korea (South), October 27 - November 2,
  2019}.\hskip 1em plus 0.5em minus 0.4em\relax {IEEE}, 2019, pp. 9626--9635.

\bibitem{RRPN}
J.~Ma, W.~Shao, Y.~Hao, W.~Li, W.~Hong, Y.~Zheng, and X.~Xue,
  ``Arbitrary-oriented scene text detection via rotation proposals,''
  \emph{IEEE Transactions on Multimedia}, vol.~PP, no.~99, p.~1, 2017.

\bibitem{r2cnn}
Y.~Jiang, X.~Zhu, X.~Wang, S.~Yang, W.~Li, H.~Wang, P.~Fu, and Z.~Luo,
  ``R$^2$cnn: Rotational region cnn for arbitrarily-oriented scene text
  detection,'' in \emph{2018 24th International Conference on Pattern
  Recognition (ICPR)}.\hskip 1em plus 0.5em minus 0.4em\relax IEEE, 2018, pp.
  3610--3615.

\bibitem{Ding.2018}
J.~Ding, N.~Xue, Y.~Long, G.~Xia, and Q.~Lu, ``Learning roi transformer for
  detecting oriented objects in aerial images,'' \emph{arXiv: Computer Vision
  and Pattern Recognition}, 2018.

\bibitem{Yang.2018}
X.~Yang, J.~Yang, Y.~Zhang, T.~Zhang, Z.~Guo, X.~Sun, and K.~Fu, ``Scrdet:
  Towards more robust detection for small, cluttered and rotated objects,'' in
  \emph{2019 {IEEE/CVF} International Conference on Computer Vision, {ICCV}
  2019, Seoul, Korea (South), October 27 - November 2, 2019}.\hskip 1em plus
  0.5em minus 0.4em\relax {IEEE}, 2019, pp. 8231--8240.

\bibitem{R3d}
X.~Yang, Q.~Liu, J.~Yan, A.~Li, Z.~Zhang, and G.~Yu, ``R3det: Refined
  single-stage detector with feature refinement for rotating object,''
  \emph{arXiv preprint arXiv:1908.05612}, 2019.

\bibitem{CSL}
X.~Yang and J.~Yan, ``Arbitrary-oriented object detection with circular smooth
  label,'' pp. 677--694, 2020.

\bibitem{DCL}
X.~Yang, L.~Hou, Y.~Zhou, W.~Wang, and J.~Yan, ``Dense label encoding for
  boundary discontinuity free rotation detection,'' pp. 15\,819--15\,829, 2021.

\bibitem{s2anet}
J.~Han, J.~Ding, J.~Li, and G.-S. Xia, ``Align deep features for oriented
  object detection,'' \emph{IEEE Transactions on Geoscience and Remote
  Sensing}, 2021.

\bibitem{middlelines}
H.~Wei, Y.~Zhang, Z.~Chang, H.~Li, H.~Wang, and X.~Sun, ``Oriented objects as
  pairs of middle lines,'' \emph{ISPRS Journal of Photogrammetry and Remote
  Sensing}, vol. 169, pp. 268--279, 2020.

\bibitem{xline}
H.~Wei, Y.~Zhang, B.~Wang, Y.~Yang, H.~Li, and H.~Wang, ``X-linenet: Detecting
  aircraft in remote sensing images by a pair of intersecting line segments,''
  \emph{IEEE Transactions on Geoscience and Remote Sensing}, 2020.

\bibitem{shi2013ship}
Z.~Shi, X.~Yu, Z.~Jiang, and B.~Li, ``Ship detection in high-resolution optical
  imagery based on anomaly detector and local shape feature,'' \emph{IEEE
  Transactions on Geoscience and Remote Sensing}, vol.~52, no.~8, pp.
  4511--4523, 2013.

\bibitem{AdaBoost}
Y.~Freund and R.~E. Schapire, ``A decision-theoretic generalization of on-line
  learning and an application to boosting,'' \emph{Journal of Computer and
  System Sciences}, vol.~55, no.~1, pp. 119--139, 1997.

\bibitem{yang2017ship}
F.~Yang, Q.~Xu, and B.~Li, ``Ship detection from optical satellite images based
  on saliency segmentation and structure-lbp feature,'' \emph{IEEE Geoscience
  and Remote Sensing Letters}, vol.~14, no.~5, pp. 602--606, 2017.

\bibitem{Liu.2017}
Z.~Liu, H.~Wang, L.~Weng, and Y.~Yang, ``Ship rotated bounding box space for
  ship extraction from high-resolution optical satellite images with complex
  backgrounds,'' \emph{IEEE Geoscience and Remote Sensing Letters}, vol.~13,
  no.~8, pp. 1074--1078, 2017.

\bibitem{Liu.2018}
Z.~Liu, J.~Hu, L.~Weng, and Y.~Yang, ``Rotated region based cnn for ship
  detection,'' in \emph{IEEE International Conference on Image Processing},
  2018.

\bibitem{cao2017realtime}
Z.~Cao, T.~Simon, S.~Wei, and Y.~Sheikh, ``Realtime multi-person 2d pose
  estimation using part affinity fields,'' in \emph{2017 {IEEE} Conference on
  Computer Vision and Pattern Recognition, {CVPR} 2017, Honolulu, HI, USA, July
  21-26, 2017}.\hskip 1em plus 0.5em minus 0.4em\relax {IEEE} Computer Society,
  2017, pp. 1302--1310.

\bibitem{ORN}
Y.~Zhou, Q.~Ye, Q.~Qiu, and J.~Jiao, ``Oriented response networks,'' in
  \emph{2017 {IEEE} Conference on Computer Vision and Pattern Recognition,
  {CVPR} 2017, Honolulu, HI, USA, July 21-26, 2017}.\hskip 1em plus 0.5em minus
  0.4em\relax {IEEE} Computer Society, 2017, pp. 4961--4970.

\bibitem{HRSC}
Z.~Liu, L.~Yuan, L.~Weng, and Y.~Yang, ``A high resolution optical satellite
  image dataset for ship recognition and some new baselines,'' in
  \emph{International Conference on Pattern Recognition Applications and
  Methods}, vol.~2.\hskip 1em plus 0.5em minus 0.4em\relax SCITEPRESS, 2017,
  pp. 324--331.

\bibitem{aod}
C.~Li, C.~Xu, Z.~Cui, D.~Wang, T.~Zhang, and J.~Yang, ``Feature-attentioned
  object detection in remote sensing imagery,'' in \emph{2019 IEEE
  International Conference on Image Processing (ICIP)}.\hskip 1em plus 0.5em
  minus 0.4em\relax IEEE, 2019, pp. 3886--3890.

\bibitem{kingma2014adam}
D.~P. Kingma and J.~Ba, ``Adam: {A} method for stochastic optimization,'' in
  \emph{3rd International Conference on Learning Representations, {ICLR} 2015,
  San Diego, CA, USA, May 7-9, 2015, Conference Track Proceedings}, Y.~Bengio
  and Y.~LeCun, Eds., 2015.

\bibitem{2017Deep}
F.~Yu, D.~Wang, E.~Shelhamer, and T.~Darrell, ``Deep layer aggregation,'' in
  \emph{2018 {IEEE} Conference on Computer Vision and Pattern Recognition,
  {CVPR} 2018, Salt Lake City, UT, USA, June 18-22, 2018}.\hskip 1em plus 0.5em
  minus 0.4em\relax {IEEE} Computer Society, 2018, pp. 2403--2412.

\bibitem{2016Stacked}
A.~Newell, K.~Yang, and J.~Deng, ``Stacked hourglass networks for human pose
  estimation,'' pp. 483--499, 2016.

\bibitem{dota}
G.~Xia, X.~Bai, J.~Ding, Z.~Zhu, S.~J. Belongie, J.~Luo, M.~Datcu, M.~Pelillo,
  and L.~Zhang, ``{DOTA:} {A} large-scale dataset for object detection in
  aerial images,'' in \emph{2018 {IEEE} Conference on Computer Vision and
  Pattern Recognition, {CVPR} 2018, Salt Lake City, UT, USA, June 18-22,
  2018}.\hskip 1em plus 0.5em minus 0.4em\relax {IEEE} Computer Society, 2018,
  pp. 3974--3983.

\bibitem{redet}
J.~Han, J.~Ding, N.~Xue, and G.-S. Xia, ``Redet: A rotation-equivariant
  detector for aerial object detection,'' in \emph{Proceedings of the IEEE/CVF
  Conference on Computer Vision and Pattern Recognition}, 2021, pp. 2786--2795.

\bibitem{Oriented}
X.~Xie, G.~Cheng, J.~Wang, X.~Yao, and J.~Han, ``Oriented r-cnn for object
  detection,'' \emph{arXiv preprint arXiv:2108.05699}, 2021.

\bibitem{BBAVectors}
J.~Yi, P.~Wu, B.~Liu, Q.~Huang, H.~Qu, and D.~Metaxas, ``Oriented object
  detection in aerial images with box boundary-aware vectors,'' in
  \emph{Proceedings of the IEEE/CVF Winter Conference on Applications of
  Computer Vision}, 2021, pp. 2150--2159.

\bibitem{DARDet}
F.~Zhang, X.~Wang, S.~Zhou, and Y.~Wang, ``Dardet: A dense anchor-free rotated
  object detector in aerial images,'' \emph{arXiv preprint arXiv:2110.01025},
  2021.

\bibitem{Gliding}
Y.~Xu, M.~Fu, Q.~Wang, Y.~Wang, K.~Chen, G.-S. Xia, and X.~Bai, ``Gliding
  vertex on the horizontal bounding box for multi-oriented object detection,''
  \emph{IEEE Transactions on Pattern Analysis and Machine Intelligence},
  vol.~43, no.~4, pp. 1452--1459, 2020.

\bibitem{DAL}
Q.~Ming, Z.~Zhou, L.~Miao, H.~Zhang, and L.~Li, ``Dynamic anchor learning for
  arbitrary-orientedd object detection,'' \emph{arXiv preprint
  arXiv:2012.04150}, vol.~1, no.~2, p.~6, 2020.

\bibitem{GWD}
X.~Yang, J.~Yan, Q.~Ming, W.~Wang, X.~Zhang, and Q.~Tian, ``Rethinking rotated
  object detection with gaussian wasserstein distance loss,'' \emph{arXiv
  preprint arXiv:2101.11952}, 2021.

\bibitem{pointbased}
K.~Fu, Z.~Chang, Y.~Zhang, and X.~Sun, ``Point-based estimator for
  arbitrary-oriented object detection in aerial images,'' \emph{IEEE
  Transactions on Geoscience and Remote Sensing}, 2020.

\bibitem{OrientedRepPoints}
W.~Li and J.~Zhu, ``Oriented reppoints for aerial object detection,''
  \emph{arXiv preprint arXiv:2105.11111}, 2021.

\bibitem{yolov3}
J.~Redmon and A.~Farhadi, ``Yolov3: An incremental improvement,'' \emph{arXiv
  preprint arXiv:1804.02767}, 2018.

\end{thebibliography}

\begin{IEEEbiography}[{\includegraphics[width=1in,height=1.25in,clip,keepaspectratio]{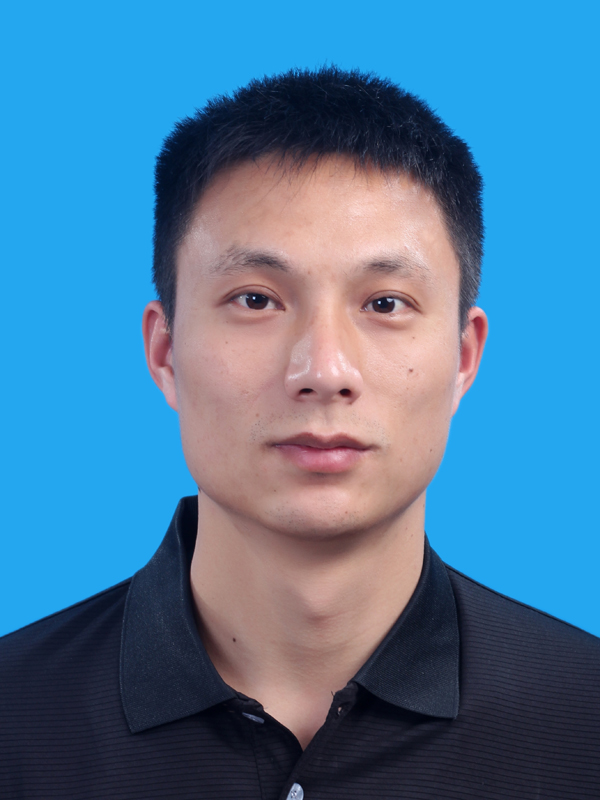}}]{Feng Zhang} received the B.E. degree in electronic information engineering from Harbin Institute of Technology(HIT), Harbin, China, in 2009, and the M.E. degree in information and communication engineering from National University of Defense Technology (NUDT),  Changsha, China, in 2011. He is currently pursuing a Ph.D. degree from the College of Electronic Science and Technology, NUDT. His research interests focus on include remote sensing image processing, pattern recognition, and computer vision.
\end{IEEEbiography}

\begin{IEEEbiography}[{\includegraphics[width=1in,height=1.25in, clip,keepaspectratio]{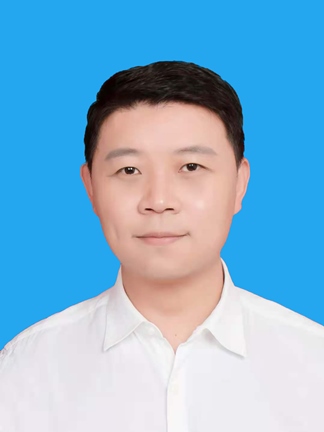}}] {Xueying Wang} received the B.S. degree in electronic information engineering from Beihang University, Beijing, China, in 2009, the M.S. and Ph.D. degrees in electronic science and technology from the National
University of Defense Technology, Changsha, China, in 2011 and 2016. He is currently an Assistant Professor with the College of Electrical Science, National University of Defense Technology. His research interests include remote sensing image processing, pattern recognition.
\end{IEEEbiography}

\begin{IEEEbiography}[{\includegraphics[width=1in,height=1.25in,clip,keepaspectratio]{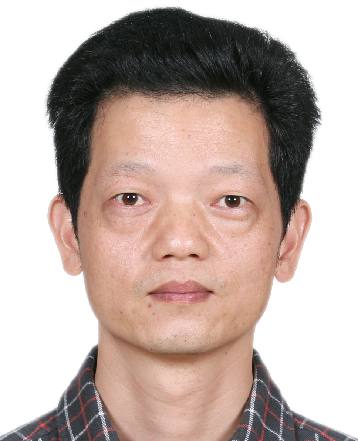}}] {Shilin Zhou} received the B.S., M.S., and Ph.D.
 degrees in electrical engineering from Hunan University, Changsha, China, in 1994, 1996, and 2000, respectively. He is currently a Full Professor with the
 College of Electrical Science, National University of Defense Technology, Changsha. He has authored or co-authored over 100 referred papers. His research interests include image processing and pattern recognition.
\end{IEEEbiography}

\begin{IEEEbiography}[{\includegraphics[width=1in,height=1.25in, clip,keepaspectratio]{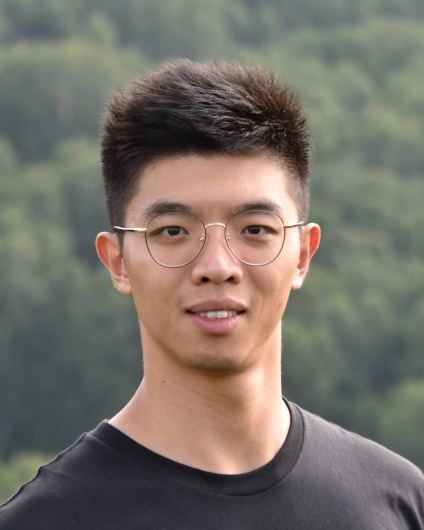}}]{Yingqian Wang} received the B.E. degree in electrical engineering from Shandong University (SDU), Jinan, China, in 2016, and the M.E. degree in information and communication engineering from National University of Defense Technology (NUDT), Changsha, China, in 2018. He is currently pursuing a Ph.D. degree from the College of Electronic Science and Technology, NUDT. He has authored several papers in journals and conferences such as TPAMI, TIP, CVPR, and ECCV. His research interests focus on low-level vision, particularly on light field imaging and image super-resolution.
\end{IEEEbiography}

\begin{IEEEbiography}[{\includegraphics[width=1in,height=1.25in,clip,keepaspectratio]{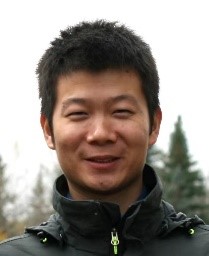}}]
  {Yi Hou} received the B.S. degree from Wuhan University, China, and the M.S. as well as a Ph.D. degree from the National University of Defense Technology, China. He held a visiting position with the Department of Computing Science, University of Alberta, Canada, from 2014 to 2016. His main research interests include robot visual SLAM, visual place recognition, time series classification, signal processing, computer vision, deep learning, pattern recognition, and image processing.
 \end{IEEEbiography}

\end{document}